\documentclass{article}

\usepackage{PRIMEarxiv}
\usepackage[utf8]{inputenc}
\usepackage[T1]{fontenc}
\usepackage{hyperref}
\usepackage{url}
\usepackage{booktabs}
\usepackage{amsfonts}
\usepackage{nicefrac}
\usepackage{microtype}
\usepackage{fancyhdr}
\usepackage{graphicx}
\graphicspath{{media/}}

\usepackage{multirow}
\usepackage{amsmath,amssymb}
\usepackage{mathrsfs}
\usepackage{xcolor}
\usepackage{textcomp}
\usepackage{algorithm}
\usepackage{algorithmic}
\usepackage{threeparttable}
\usepackage{rotating}
\usepackage{setspace}
\usepackage{tabularx}

\title{Knowledge-Guided Brain Tumor Segmentation via Synchronized Visual-Semantic-Topological Prior Fusion
\thanks{\textit{\underline{Citation}}: 
\textbf{Zhang, M.-D., Pan, K.-W. Knowledge-Guided Brain Tumor Segmentation via Synchronized Visual-Semantic-Topological Prior Fusion. Submitted to arXiv, 2025.}}
}

\author{
  Ming-Da Zhang, Kai-Wen Pan \\
  School of Software \\
  Yunnan University \\
  Kunming, Yunnan Province, China\\
  \texttt{yao110002@gmail.com} \\
}

\begin{document}
\maketitle

\begin{abstract}
\textbf{Background:} Brain tumor segmentation requires precise delineation of hierarchical structures from multi-sequence MRI. However, existing deep learning methods primarily rely on visual features, showing insufficient discriminative power in ambiguous boundary regions. Moreover, they lack explicit integration of medical domain knowledge such as anatomical semantics and geometric topology.

\textbf{Methods:} We propose a knowledge-guided framework, Synchronized Tri-modal Prior Fusion (STPF), that explicitly integrates three heterogeneous knowledge priors: pathology-driven differential features (T1ce-T1, T2-FLAIR, T1/T2) encoding contrast patterns, unsupervised semantic descriptions transformed into voxel-level guidance via spatialization operators, and geometric constraints extracted through persistent homology analysis. A dual-level fusion architecture dynamically allocates prior weights at the voxel level based on confidence and at the sample level through hypernetwork-generated conditional vectors. Furthermore, nested output heads structurally ensure the hierarchical constraint ET$\subseteq$TC$\subseteq$WT.

\textbf{Results:} STPF achieves a mean Dice coefficient of 0.868 on the BraTS 2020 dataset, surpassing the best baseline by 2.6 percentage points (3.09\% relative improvement). Notably, five-fold cross-validation yields coefficients of variation between 0.23\% and 0.33\%, demonstrating stable performance. Additionally, ablation experiments show that removing topological and semantic priors leads to performance degradation of 2.8\% and 3.5\%, respectively.

\textbf{Conclusions:} By explicitly integrating medical knowledge priors—anatomical semantics and geometric constraints—STPF improves segmentation accuracy in ambiguous boundary regions while demonstrating generalization capability and clinical deployment potential.
\end{abstract}


\section{Introduction}\label{sec:intro}

Gliomas account for 80\% of primary adult brain tumors\cite{aiinglioma2025,eurradiol2023gbm}. Notably, high-grade glioma patients face a poor prognosis, with median survival of less than two years and five-year survival rates below 10\%. Consequently, precise tumor segmentation is important for personalized treatment, directly impacting critical clinical decisions such as surgical planning, radiation therapy target delineation, and treatment efficacy assessment\cite{frontiersglioma2022,frontiersposttx2022}. To achieve this precision, multi-sequence MRI provides complementary tissue contrast information, thereby establishing a multi-modal data foundation for fine delineation of tumor sub-regions\cite{noa2024datasets}. However, the tissue heterogeneity, irregular morphology, and ambiguous boundaries of brain tumors, combined with the mandatory nested hierarchical relationship between enhancing tumor (ET), tumor core (TC), and whole tumor (WT) (ET$\subseteq$TC$\subseteq$WT), make automatic segmentation a major challenge in computational neuro-oncology. In response to this challenge, the Brain Tumor Segmentation (BraTS) challenge series\cite{brats2021,bonato2025,bratslighthouse2025} has continuously advanced this field, expanding from initial glioma segmentation to meningioma\cite{bratsmenrt2025}, brain metastases\cite{bratsmets2025}, and pathology image analysis\cite{bratspath2024}.

Deep learning methods have achieved progress in brain tumor segmentation. In particular, advanced architectures represented by nnU-Net\cite{isensee2021nnunet} and Swin UNETR\cite{hatamizadeh2022swinunetr} have reached expert-level performance on BraTS challenges through adaptive configuration and global context modeling. Recent methods include Mamba-based long-range sequence modeling\cite{xing2024segmamba,xing2025segmambav2}, diffusion model enhancement\cite{xing2025diffunet,ding2024fdifffusion}, hybrid masked modeling\cite{xing2024hybridmim}, and hierarchical context interaction\cite{hciwacv2025}. Nevertheless, these methods focus primarily on architectural innovations in visual feature space and represent essentially data-driven visual learning. Consequently, they exhibit three systematic limitations: First, insufficient discriminative capability in regions with ambiguous boundaries and unreliable signal intensity, as highlighted by HD distance evaluation studies\cite{hdilemma2024} and the Metrics Reloaded framework\cite{metricsreloaded2024}. Second, network outputs may contain topologically unreasonable structures, thereby producing fatal defects in downstream clinical applications. Third, insufficient utilization of medical domain knowledge—clinicians comprehensively consider semantic information such as anatomical location, morphological features, and growth patterns during diagnosis, as well as geometric constraints that tumors spread along specific pathways while maintaining particular connectivity. These clinical reasoning processes embody knowledge priors that remain underexploited in current data-driven approaches.

The development of vision-language models (VLMs) provides new pathways for integrating semantic knowledge priors. For instance, foundation models like BiomedCLIP\cite{biomedclip2025} demonstrate zero-shot and few-shot learning capabilities through large-scale image-text pair pretraining. Similarly, BioViL-T\cite{biovilt2022} proves that even limited medical text can improve visual representation learning. However, integrating semantic knowledge into dense prediction tasks still faces three obstacles: First, medical datasets generally lack high-quality text annotations, thus requiring unsupervised semantic generation pipelines. Second, medical semantics require precise anatomical and pathological terminology rather than vague descriptions. Third, existing methods treat semantic embeddings as global conditional signals, thereby lacking mechanisms to map global descriptions to voxel-level spatial distributions. Attempts by ConTEXTual Net\cite{contextualnet2023} and SEG-SAM\cite{segsam2024} indicate that semantic guidance can improve segmentation performance; nevertheless, more comprehensive multi-modal prior fusion is still needed. Similarly, while persistent homology theory\cite{topoloss2022,toporeview2025} provides tools for capturing geometric structure priors, how to transform abstract topological features into collaboratively optimizable spatial representations remains an open question.

To address these challenges, we propose a knowledge-guided framework, Synchronized Tri-modal Prior Fusion (STPF), which achieves synchronized interaction of three heterogeneous knowledge priors—visual, semantic, and topological—at each decoder layer. Note that the "tri-modal prior" in this paper refers to three types of heterogeneous knowledge priors (visual pathology patterns, semantic anatomical descriptions, topological geometric constraints), while the input data consists of four-sequence MRI (T1, T1ce, T2, FLAIR). Core contributions include:

(1) Knowledge-driven multi-modal prior generation and spatialization: We explicitly construct pathology-driven differential features as visual knowledge priors. Additionally, we generate standardized clinical descriptions through unsupervised anomaly detection as semantic knowledge priors. Moreover, we extract geometric constraints using persistent homology as topological knowledge priors, and transform all three-way priors into voxel-level distributions through spatialization operators.

(2) Dual-level adaptive prior fusion architecture: At the voxel level, we dynamically allocate tri-modal prior weights based on confidence. Furthermore, at the sample level, we uniformly adjust decoder features through hypernetwork-generated modulation parameters.

(3) Unified prior fusion and structured output: We transform heterogeneous knowledge priors into logit space for energy superposition decision-making. Simultaneously, we structurally guarantee the hierarchical relationship ET$\subseteq$TC$\subseteq$WT through nested output heads.

\section{Related Work}\label{sec:related}

\subsection{Deep Learning for Brain Tumor Segmentation}

Since its launch in 2012, the BraTS challenge has established standardized evaluation protocols for brain tumor segmentation, defining three hierarchical regions: whole tumor (WT), tumor core (TC), and enhancing tumor (ET)\cite{brats2021}. Early methods, represented by 3D U-Net\cite{cicek20163d}, capture multi-scale context through encoder-decoder structures. Subsequently, the nnU-Net\cite{isensee2021nnunet} proposed by Isensee et al. achieves a mean Dice coefficient of 0.850 on BraTS 2020 through adaptive preprocessing and dynamic configuration. Following this, Myronenko\cite{myronenko2019autoencoder} introduces autoencoder regularization, while SegResNet\cite{segresnetmonai2020} enhances gradient flow through residual connections. Furthermore, inspired by Transformer success in natural language processing, Hatamizadeh et al. propose UNETR\cite{hatamizadeh2022unetr}, introducing pure Transformer encoders to 3D segmentation. Building upon this, Tang et al.'s Swin UNETR\cite{hatamizadeh2022swinunetr} achieves 0.856 on BraTS 2021 through hierarchical Shifted Window mechanisms. Meanwhile, TransBTS\cite{wang2021transbts} and TransUNet\cite{chen2024transunet} explore hybrid CNN-Transformer architectures, and SwinBTS\cite{swinbts2022} further optimizes window attention mechanisms.

Recent work exhibits more diverse innovation paths. For example, CKD-TransBTS\cite{lin2023ckdtransbts} introduces cross-knowledge distillation, while FDiff-Fusion\cite{ding2024fdifffusion} combines diffusion models and fuzzy learning. Moreover, the SegMamba series\cite{xing2024segmamba,xing2025segmambav2} achieves efficient long-range modeling through state space models, and Diff-UNet\cite{xing2025diffunet} embeds diffusion processes into segmentation frameworks. Although these methods achieve progress in performance metrics, they mainly focus on architectural-level innovations with insufficient integration of medical domain knowledge priors. Consequently, these pure data-driven methods show decreased discriminative capability in ambiguous boundary regions, optimize through soft loss functions, and lack explicit constraints on topological rationality and hierarchical consistency. Semi-supervised and weakly-supervised methods\cite{chen2023semisupervised,chen2024dynamiccontrastive,chen2025crossimagematching,chen2020affinityguided,chen2025fromgaze} demonstrate potential in data-limited scenarios but still require stronger prior constraints.

\subsection{Semantic Knowledge Integration}

The development of vision-language models introduces a semantic understanding dimension to medical image analysis. Specifically, CLIP\cite{radford2021clip} demonstrates zero-shot classification capabilities through contrastive learning on 400 million image-text pairs. Building on this foundation, BiomedCLIP\cite{biomedclip2025} is pretrained on 15 million biomedical image-text pairs from PubMed Central, specifically optimized for medical terminology. As a result, it outperforms general models in report generation and disease classification tasks. Similarly, BioViL-T\cite{biovilt2022} proves that even limited medical text can improve visual representations. Regarding practical applications, ConTEXTual Net\cite{contextualnet2023} is pioneering work introducing textual information into segmentation tasks, combining U-Net and T5-Large language models, fusing text and visual features through cross-attention, and achieving a Dice coefficient of 0.716 in pneumothorax segmentation. Likewise, SEG-SAM\cite{segsam2024} integrates LLM-generated descriptions into the SAM framework.

However, integrating semantic knowledge priors into dense prediction tasks still faces three challenges: First, mainstream datasets lack text annotations and lack end-to-end automation pipelines. Second, medical semantics require precise anatomical and pathological terminology rather than vague descriptions. Third, existing methods treat semantic embeddings as global conditional signals, thereby lacking mapping mechanisms for voxel-wise spatial distributions. Attempts by ConTEXTual Net and SEG-SAM indicate that semantic guidance can improve segmentation performance; however, the granularity gap between global semantic descriptions and dense prediction tasks remains unresolved. The spatialization operator proposed in this paper transforms global semantic tokens into spatial representations at the same resolution through probability maps and distance field transforms, thereby enabling semantic knowledge priors to function at each voxel location.

\subsection{Topological and Geometric Constraints}

Topological data analysis provides geometric constraint tools for medical image segmentation. Specifically, persistent homology theory\cite{topoloss2022} captures multi-scale topological features of data by analyzing connectivity changes at different thresholds. The topological loss function proposed by Clough et al.\cite{topoloss2022} constrains network output connectivity through Betti numbers, reducing unreasonable isolated regions in cardiac segmentation tasks. Furthermore, a review of topological descriptors in medical imaging\cite{toporeview2025} indicates that geometric knowledge priors are important for improving anatomical rationality of segmentation. In addition, graph neural networks\cite{gnnmedsurvey2023} provide another approach to modeling spatial relationships, with the multi-class graph reasoning method at MICCAI 2024\cite{miccai2024gnnbrain} demonstrating GNN potential in brain tumor segmentation.

However, existing methods mainly apply topological constraints as post-processing steps or loss function terms, thereby lacking deep fusion with visual features. Transforming abstract topological knowledge priors into collaboratively optimizable spatial representations remains an open question. This paper processes skeleton graphs extracted by persistent homology through graph attention networks\cite{gat} and designs spatialization operators to map abstract graph structures to dense spatial representations, thus achieving voxel-wise interaction between topological priors and visual features. Meanwhile, unsupervised anomaly detection methods provide a technical foundation for the semantic generation module in this paper. ASC-Net\cite{ascnetmiccai2021} achieves unsupervised anomaly segmentation through adversarial training, Mahalanobis distance methods\cite{miccai2024mahauad} detect brain anomalies using statistical distributions, and diffusion model-based contrastive analysis\cite{uaddiffusion2025} demonstrates generative model potential in anomaly detection, thereby enabling the system to generate standardized clinical descriptions without manual annotation.

\subsection{Multi-modal MRI Feature Fusion}

Effective utilization of multi-sequence MRI is key to precise brain tumor segmentation. Different sequences exhibit differential sensitivity to pathological tissues: T1-weighted provides anatomical structure, T1 contrast-enhanced highlights blood-brain barrier disruption regions, T2-weighted detects edema and cystic changes, and FLAIR sequence separates peritumoral edema by suppressing cerebrospinal fluid signals. Early fusion strategies concatenate multi-sequences as multi-channel inputs, implicitly assuming networks automatically learn cross-modal relationships; however, simultaneously learning intra-modal features and inter-modal relationships increases optimization difficulty. To address this, MAMC\cite{mamcattn2021} dynamically weights different sequences through channel attention, NestedFormer\cite{xing2022nestedformer} introduces modality-sensitive gating to implement cross-modal feature propagation at different scales, and CMAF-Net\cite{cmafnet2024} designs cross-modal attention fusion mechanisms.

However, these methods treat each sequence as an independent information source, insufficiently considering the pathological knowledge embedded in inter-sequence differential features. Medical imaging indicates that inter-sequence contrast patterns directly correspond to specific pathological features: T1ce-T1 difference highlights contrast agent leakage regions, T2-FLAIR difference separates free water and bound water to delineate whole tumor boundaries, and T1/T2 ratio enhances necrotic tissue contrast. Furthermore, existing fusion strategies treat different modalities as equal partners, ignoring essential differences in prior representations: visual modality is dense spatial features, semantic modality is sparse global descriptions, and topological modality is abstract graph structures. The dual-level fusion architecture proposed in this paper dynamically allocates fusion weights through confidence estimation, thereby adaptively relying on the most reliable knowledge prior in different spatial locations and pathological scenarios.

\section{Method}\label{sec:method}

\begin{figure}[htbp]
\centering
\includegraphics[width=\textwidth]{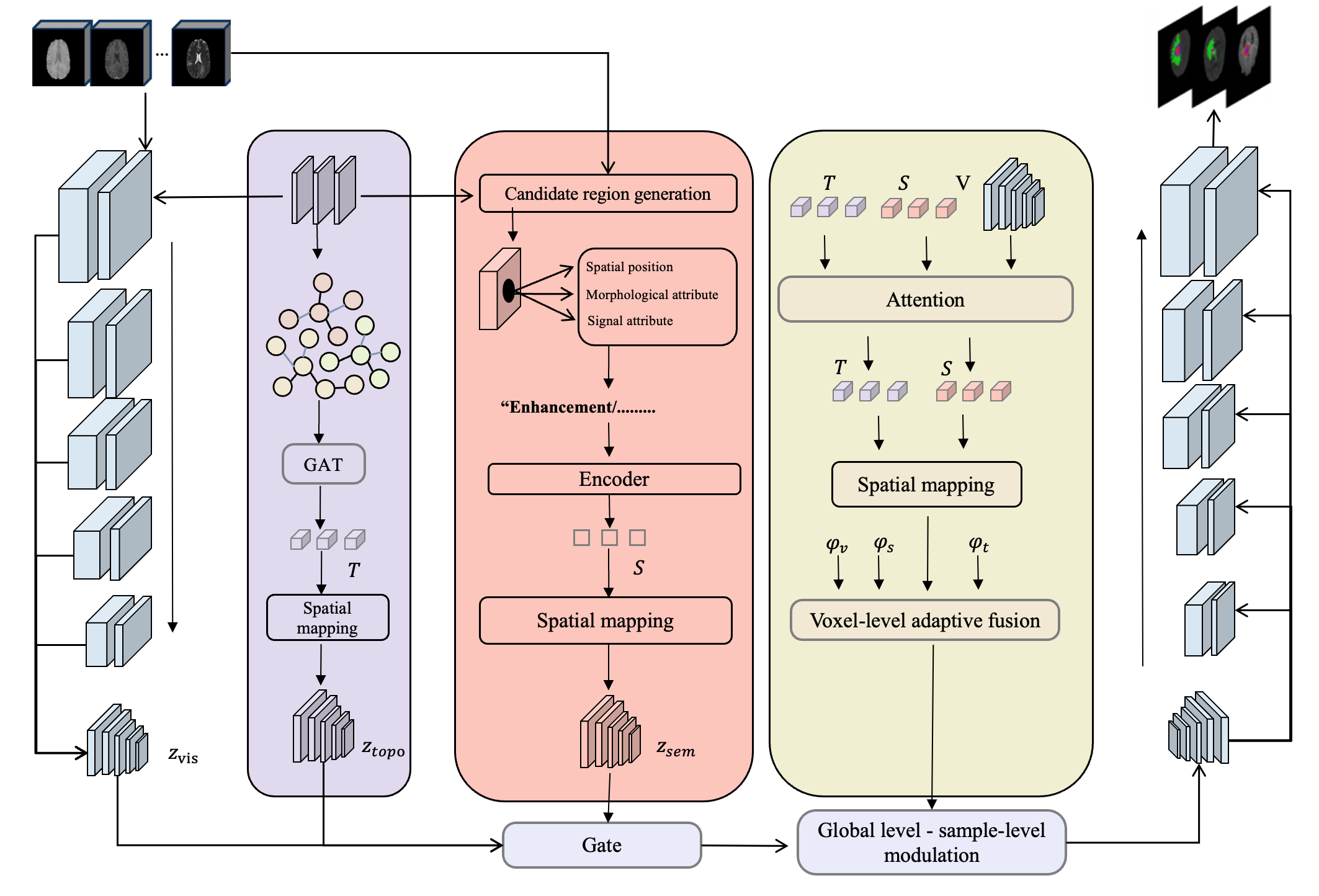}
\caption{STPF framework diagram. The left encoder extracts visual features from multi-sequence MRI, obtaining \texttt{z\_vis} at the bottom. The purple topological path generates topological tokens $T$ via GAT and obtains \texttt{z\_topo} through spatialization. The pink semantic path includes candidate region generation and attribute extraction, obtaining semantic tokens $S$ through Encoder and \texttt{z\_sem} through spatialization. Three-way knowledge priors are fused at the bottom through Gate and sent to global level–sample-level modulation. The yellow-green section performs Attention on $T$, $S$, $V$, then completes voxel-level fusion through spatialization and voxel-level adaptive fusion (estimating weights via $\phi_v,\ \phi_s,\ \phi_t$), progressively restoring on the right side and outputting segmentation results.}
\label{fig:overview}
\end{figure}

This paper proposes a knowledge-guided brain tumor segmentation framework based on synchronized tri-modal prior fusion (STPF), which performs synchronized collaborative decision-making on image-enhanced features, semantic descriptions, and topological constraints at each decoder layer. As shown in Figure~\ref{fig:overview}, the method starts from four-sequence MRI data and extracts knowledge priors through three parallel paths. The visual path constructs a multi-scale feature pyramid based on multi-modal difference maps encoding pathological knowledge. Simultaneously, the semantic path generates structured descriptions through anomaly detection, representing anatomical knowledge. In parallel, the topological path extracts geometric constraints using persistent homology analysis, capturing morphological knowledge. After fusing three-way priors into sample-level conditional vectors, they are input to a hypernetwork to generate modulation parameters. The decoder implements dual-level prior fusion: at the global level through sample-level modulation to unify feature representations, and at the local level dynamically allocating fusion weights based on confidence at each voxel location. Finally, the unified prior fusion module transforms all knowledge priors into the same logit space for negotiation, coordinated with nested output heads ensuring hierarchical inclusion relationships, employing stick-breaking parameterization\cite{stickbreakingvb2020}.

\subsection{Multi-modal Prior Generation}

\subsubsection{Visual Enhancement Modality: Multi-modal Differential Feature Construction as Pathological Knowledge Prior}

In medical image analysis, contrast information between different MRI sequences contains lesion features representing pathological knowledge. Based on standard four-sequence MRI (T1, T1ce, T2, FLAIR), this paper explicitly constructs inter-sequence differential features to encode domain-specific pathological patterns. To capture features from different regions, we define three differential channels as follows:

\begin{align}
D_{\text{enhance}} &= \frac{T1_{ce} - T1}{T1 + \epsilon} \quad \text{(Enhancement difference)} \\
D_{\text{edema}} &= T2 - \text{FLAIR} \quad \text{(Edema difference)} \\
D_{\text{necrosis}} &= \frac{T1}{T2 + \epsilon} \quad \text{(Necrosis contrast)}
\end{align}

Specifically, the enhancement difference captures contrast-enhanced regions corresponding to enhancing tumor (ET). Meanwhile, the edema difference uses T2 and FLAIR signal differences to separate edema regions, thereby locating whole tumor (WT) extent. Furthermore, the necrosis contrast utilizes T1 and T2 complementarity to highlight necrotic regions, thus assisting tumor core (TC) segmentation. Collectively, these differential features explicitly encode radiological knowledge about tumor pathology.

Following logarithmic compression and quantile clipping, difference maps are normalized to the $[0,1]$ interval, thereby expanding the original four channels to a seven-channel enhanced representation. Subsequently, a dual-branch encoder processes the original four-sequence channels and differential three channels separately: the original branch captures basic anatomical structure and signal intensity information, while the differential branch focuses on pathological contrast patterns. The two branches interact and fuse through cross-attention mechanisms at each scale, progressively downsampling to extract a multi-scale feature pyramid $\{V_\ell\}_{\ell=1}^L$, which provides visual feature representations for subsequent decoder layers.

\subsubsection{Topological Modality: Geometric Structure Prior Extraction as Morphological Knowledge}

Visual and semantic modalities provide texture features and region attributes respectively; however, they lack explicit constraints on target geometric morphology. To address this limitation, the topological modality analyzes regional geometric structures through persistent homology\cite{topoloss2022,toporeview2025}, extracting geometric knowledge priors that encode how tumors grow and spread. The system defines field functions based on the three differential channels: $S_{\text{ET}}$ takes high values in enhanced non-necrotic regions, $S_{\text{TC}}$ captures both enhanced and necrotic signals, and $S_{\text{WT}}$ is primarily dominated by edema signals. Subsequently, topological analysis is performed for each field, retaining stable connected regions while suppressing unstable isolated points and noise. Following this analysis, skeletons are extracted for stable regions and graph structures $G_c = (V_c, E_c)$ are constructed, where nodes contain key points and their attributes (field values, stability, local features), and edges record inter-region connections and geometric attributes. Consequently, graph attention networks (GAT)\cite{gat,gnnmedsurvey2023,miccai2024gnnbrain} jointly process three graph structures to extract node embeddings:

\begin{equation}
T = \{t_j\}_{j=1}^{K_t} = \text{GAT}(G_{\text{ET}} \cup G_{\text{TC}} \cup G_{\text{WT}})
\end{equation}

obtaining $K_t$ topological tokens. These abstract graph embeddings need to be transformed into spatial representations to interact with visual features. Accordingly, the spatialization operator $\Pi_T$ maps graph structures back to voxel space:

\begin{equation}
P_T^{(\ell)}(x) = \Pi_T(T, G_c) = \sum_{e \in E_c} \exp\left(-\frac{d(x, e)^2}{2r_e^2}\right) \cdot t_e
\end{equation}

For each graph edge $e$, a Gaussian response region is constructed centered on its skeleton with radius $r_e$ as the scale. For each node, a radial basis function is constructed. The distance $d(x, e)$ from each voxel $x$ to edge $e$ determines the influence strength of that edge on that voxel, thereby transforming the abstract graph structure into a spatial field that can directly interact with visual features. Consequently, this enables geometric knowledge constraints of tumor growth along the skeleton to function at each voxel location.

\subsubsection{Semantic Modality: Structured Region Description and Spatialization as Anatomical Knowledge}

While visual features can capture texture and intensity patterns, they lack explicit representation of region location and attributes. To address this deficiency, the semantic modality transforms imaging space into structured natural language descriptions, thereby encoding anatomical knowledge priors. Since the BraTS dataset does not include text annotations, this paper designs an automated generation pipeline from images to text, inspired by unsupervised anomaly detection methods\cite{ascnetmiccai2021,miccai2024mahauad,uaddiffusion2025}. This pipeline includes three key stages, as shown in Algorithm~\ref{alg:semantic}: first detecting candidate regions through unsupervised methods, then extracting multi-dimensional attributes, and finally assembling standardized text descriptions.

\begin{algorithm}[h]
\caption{Image-to-semantic description generation pipeline}
\label{alg:semantic}
\begin{algorithmic}[1]
\REQUIRE Four-sequence MRI: $\{I_m\}_{m \in \{\text{T1, T1ce, T2, FLAIR}\}}$
\ENSURE Semantic token sequence $S = \{s_k\}_{k=1}^{K_s}$
\STATE \textbf{Stage 1: Candidate region detection}
\STATE Train denoising autoencoders for each modality, then compute reconstruction error maps
\STATE In parallel, compute left-right hemisphere symmetry difference maps
\STATE Subsequently, fuse two-way anomaly responses to obtain candidate region set $\{R_i\}_{i=1}^{N_r}$
\STATE \textbf{Stage 2: Attribute extraction (encoding anatomical knowledge)}
\FOR{each candidate region $R_i$}
    \STATE First, extract spatial attributes: centroid coordinates, volume, brain region location (mapped via Automated Anatomical Labeling 3 (AAL3) standard atlas\cite{aal3hbm2020})
    \STATE Next, extract morphological attributes: compactness, principal axis direction, shape irregularity
    \STATE Finally, extract signal attributes: statistics for each modality and differential channel responses
\ENDFOR
\STATE \textbf{Stage 3: Text assembly and encoding}
\STATE Discretize continuous attributes into controlled vocabulary labels (see Table~\ref{tab:vocab})
\STATE Then, assemble natural language descriptions according to templates
\STATE Parse into JSON structured data: discrete fields $\oplus$ continuous fields
\STATE Finally, $S \leftarrow \text{MLP}_{\text{sem}}(\text{Concat}(\text{Embed}(\text{discrete}), \text{Norm}(\text{continuous})))$
\RETURN Semantic token sequence $S$
\end{algorithmic}
\end{algorithm}

To ensure medical standardization and semantic consistency of generated text, Stage 3 of the algorithm employs a controlled vocabulary mechanism. Table~\ref{tab:vocab} shows standardized vocabularies for each attribute type, with location terms referenced from AAL3 atlas\cite{aal3hbm2020} definitions ensuring anatomical accuracy. Moreover, volume, morphological, and signal features are all discretized into fixed levels, thereby avoiding synonym confusion.

\begin{table}[h]
\centering
\caption{Controlled vocabulary examples (encoding anatomical knowledge)}
\label{tab:vocab}
\begin{tabularx}{\textwidth}{lX}
\hline
\textbf{Attribute Type} & \textbf{Vocabulary} \\
\hline
Laterality & Left, Right \\
Location & Frontal lobe, Temporal lobe, Parietal lobe, Occipital lobe, Basal ganglia, Thalamus, ... (ref. AAL3 atlas) \\
Volume level & Small, Medium, Large \\
Morphology & Regular spherical, Ellipsoidal, Irregular \\
Enhancement pattern & Prominent, Moderate, Mild \\
Edema extent & Extensive, Moderate, Localized \\
\hline
\end{tabularx}
\end{table}

The algorithm and vocabulary jointly achieve unsupervised semantic generation: locating anomalous regions through reconstruction error and symmetry analysis, extracting spatial, morphological, and signal attributes then mapping to discrete labels, and finally assembling standardized descriptions. Notably, this pipeline requires no manual annotation while ensuring both medical accuracy and facilitating subsequent network encoding.

Generated natural language descriptions need to be further transformed into spatial distributions that can interact with visual features. This transformation process consists of two steps: text encoding and spatial mapping. In the encoding stage, the system first parses descriptions into JSON structured data, where discrete fields (such as laterality, location, morphology) are mapped to dense vectors through embedding layers, continuous fields (such as precise volume values, compactness) undergo z-score normalization, and after concatenation are input to a multi-layer perceptron (MLP) to obtain $K_s$ semantic token sequences of fixed dimension:
\begin{equation}
S = \{s_k\}_{k=1}^{K_s} = \text{MLP}_{\text{sem}}(\text{Concat}(\text{Embed}(\text{discrete}), \text{Norm}(\text{continuous})))
\end{equation}

After obtaining global semantic tokens, the spatialization operator $\Pi_S$ maps them to spatial distributions at the same resolution as visual features. This operator calculates weights at each voxel location based on region probability map $P_{\text{anat}}$ and distance field $D_{\text{anat}}$, thereby achieving spatial localization of semantic attributes:
\begin{equation}
P_S^{(\ell)}(x) = \Pi_S(S, P_{\text{anat}}, D_{\text{anat}}) = \sum_k w_k(x) \cdot s_k
\end{equation}
Weight $w_k(x)$ enables strong activation of corresponding semantic tokens for voxels within specific regions, while weights decay for voxels far from that region. Consequently, the semantic encoder is initialized based on Transformers pretrained on medical text corpora, with hierarchical learning rate strategies during fine-tuning and data augmentation (random dropping of non-critical attributes, synonym replacement, etc.) to enhance robustness. Semantic priors thus transform from global anatomical knowledge vectors into distributional representations that can dynamically interact with visual features at each spatial location.

The three paths complete transformation from raw MRI to structured knowledge priors: the visual modality generates multi-scale feature pyramid $\{V_\ell\}$ encoding pathological patterns, the semantic modality generates tokens $S$ and their spatialization $P_S^{(\ell)}$ encoding anatomical knowledge, and the topological modality generates graph tokens $T$ and their spatialization $P_T^{(\ell)}$ encoding geometric constraints. To achieve sample-level global guidance, the system fuses global representations of three-way priors into a unified conditional vector through adaptive gating. First, global pooling is performed on three-way priors separately, then contribution degrees of each modality are dynamically allocated through attention mechanisms:

\begin{equation}
z = \sum_i \alpha_i \mathbf{z}_i, \quad \alpha_i = \text{softmax}(w_i^\top [\mathbf{z}_{\text{vis}} \| \mathbf{z}_{\text{sem}} \| \mathbf{z}_{\text{topo}}])
\end{equation}

where $\mathbf{z}_{\text{vis}}, \mathbf{z}_{\text{sem}}, \mathbf{z}_{\text{topo}}$ represent global vector representations of visual, semantic, and topological paths respectively. This conditional vector encodes global characteristics of the current sample and is input to a lightweight hypernetwork to generate modulation parameters: channel-level affine coefficients $\gamma(z), \beta(z)$ for feature modulation, and depthwise separable convolution kernels $K(z)$ for adaptive filtering. Consequently, this enables the network to adjust feature extraction and fusion strategies based on sample characteristics (such as location, morphology, size).

\subsection{Dual-Level Fusion Architecture}

The prior generation stage provides multi-modal knowledge information for subsequent fusion. To effectively utilize these priors, this method employs a dual-level fusion architecture: at the local level, three knowledge priors undergo confidence-weighted synchronized fusion at each voxel location, thereby achieving dynamic balance of visual features, semantic constraints, and topological priors. Simultaneously, at the global level, sample-level conditional vectors generate modulation parameters for unified adaptive adjustment of fused features.

\subsubsection{Decoder Layer-wise Fusion}

Each upsampling layer $\ell$ of the decoder performs dual-level fusion. Inputs include visual features $V_\ell$ from the encoder, semantic tokens $S$ and their spatialization $P_S^{(\ell)}$, topological tokens $T$ and their spatialization $P_T^{(\ell)}$, and sample-level conditional vector $z$.

The fusion process first establishes information exchange channels among the three knowledge priors. Visual features $V_\ell$ as Query separately query semantic tokens $S$ and topological tokens $T$, obtaining guidance from region attributes and geometric constraints. Simultaneously, semantic and topological tokens as Query query visual features, absorbing visual evidence from the current layer and updating their own representations:

\begin{equation}
S' = S + \text{Attention}(S, V_\ell), \quad T' = T + \text{Attention}(T, V_\ell)
\end{equation}

This bidirectional interaction enables knowledge priors to be refined layer by layer during decoding—shallow layers based on coarse candidate regions, deep layers incorporating more visual details. Updated $S', T'$ generate distributions for the current layer through spatialization operators, thereby providing same-resolution modal representations for subsequent fusion.

After obtaining same-resolution tri-modal representations, the system calculates fusion weights at each voxel location $x$. Confidence of three-way features is estimated through lightweight convolution $\phi_*$ and normalized via softmax:

\begin{align}
[\alpha_v(x), \alpha_s(x), \alpha_t(x)] &= \text{softmax}\big(\phi_v(V_\ell(x)), \phi_s(P_S^{(\ell)}(x)), \phi_t(P_T^{(\ell)}(x))\big) \\
F_\ell(x) &= \alpha_v(x) \cdot V_\ell(x) + \alpha_s(x) \cdot P_S^{(\ell)}(x) + \alpha_t(x) \cdot P_T^{(\ell)}(x)
\end{align}

where $\phi_*$ combines feature content with data quality (signal-to-noise ratio, reconstruction error) to output confidence. The three-way weights satisfy $\sum \alpha = 1$, thereby achieving adaptive fusion based on confidence of each knowledge prior at that location. For example, semantic weight $\alpha_s$ is higher when regional structure is clear, topological weight $\alpha_t$ is higher when boundaries are ambiguous but skeleton is stable, and visual weight $\alpha_v$ is higher when signal quality is good.

After obtaining fused features $F_\ell$, the system further applies sample-level adaptive adjustment. Conditional vector $z$ generates two types of modulation parameters through a lightweight hypernetwork: channel-level affine coefficients $\gamma(z), \beta(z)$ apply uniform transformation to all channels; for the differential branch, additional depthwise separable convolution kernels $K(z)$ perform adaptive filtering, with this filter acting on subspaces of differential-related channels through channel grouping mechanisms:

\begin{equation}
F_\ell \leftarrow \gamma(z) \odot F_\ell + \beta(z), \quad F_{\text{diff}} \leftarrow F_{\text{diff}} + \text{DepthwiseConv}(F_{\text{diff}}; K(z))
\end{equation}

This global guidance ensures that all spatial locations of the same sample share a unified adjustment tendency. After skip connections and residual block transmission, all decoder layers $\ell=L, \ldots, 1$ progressively restore resolution, finally obtaining high-resolution features $F_1$.

\subsubsection{Unified Prior Fusion Module}

The high-resolution features $F_1$ at the decoder top layer have fused visual representations from all levels; however, semantic and topological priors still need to play a role in final classification decisions. To address this, this paper designs a Unified Prior Fusion (UPF) module to transform all knowledge priors into a unified logit space. UPF has clear division of responsibilities: base logits carry visual evidence accumulated from all fusion layers, while prior energy terms specifically handle semantic and topological constraints, with both achieving decision negotiation through energy addition.

For each class $c \in \{\text{WT}, \text{TC}, \text{ET}\}$, UPF generates prior energy through MLP:

\begin{align}
\psi_c(x) &= \text{MLP}\big([x, z, P_S^{(1)}(x), P_T^{(1)}(x)]\big) \\
\text{logit}_c(x) &= \text{logit}_c^{\text{base}}(x) + \psi_c(x)
\end{align}

where $\text{logit}_c^{\text{base}}$ is the base logit obtained from final features $F_1$ through a classification head, encoding visual evidence accumulated from all fusion layers. $x$ is spatial coordinates, $z$ is sample-level conditional vector, and $P_S^{(1)}, P_T^{(1)}$ are semantic and topological spatialization features of the final layer. The MLP maps these inputs to prior energy $\psi_c(x)$, representing the likelihood that location $x$ belongs to class $c$ based on current knowledge priors.

Final classification logits adopt energy addition for decision-making. When visual evidence is strong and priors are consistent, energy superposition produces high confidence. When visual uncertainty exists but priors are explicit, prior energy compensates for visual insufficiency. When priors conflict but visual discriminability is strong, base logits dominate. To prevent prior energy from overly dominating decisions under low confidence, magnitude regularization is applied to $\psi_c$ during training (increasing L2 penalty when prior confidence is below threshold), while introducing prior alignment terms in high-confidence regions to gently draw predictions closer to prior soft labels.

\subsubsection{Nested Output Head}

Final classification logits need to be converted to probability distributions while ensuring hierarchical inclusion relationships $\text{ET} \subseteq \text{TC} \subseteq \text{WT}$. Traditional methods penalize predictions violating hierarchy through loss functions, but this is only a soft constraint—networks may still output illegal results. To structurally guarantee output legality, stick-breaking parameterization\cite{stickbreakingvb2020} embeds hierarchical relationships into the network:

\begin{align}
p_{\text{WT}}(x) &= \sigma(\theta_1(x)) \\
p_{\text{TC}}(x) &= p_{\text{WT}}(x) \cdot \sigma(\theta_2(x)) \\
p_{\text{ET}}(x) &= p_{\text{TC}}(x) \cdot \sigma(\theta_3(x))
\end{align}

where $\theta_i(x)$ are intermediate parameters obtained from final logits through affine projection. This parameterization ensures: regardless of $\theta_i$ values, output probabilities necessarily satisfy $p_{\text{ET}} \leq p_{\text{TC}} \leq p_{\text{WT}}$. Each subset's probability is expressed as its parent set probability multiplied by conditional probability: $p_{\text{TC}}$ represents the probability of belonging to TC given that it is WT, and $p_{\text{ET}}$ represents the probability of belonging to ET given that it is TC.

\subsection{Loss Function Design}

Model training adopts a multi-objective joint optimization strategy, with the overall loss function composed of four complementary parts, respectively constraining segmentation accuracy, hierarchical consistency, spatial continuity, and topological rationality. The total loss function is defined as:

\begin{equation}
\mathcal{L} = \mathcal{L}_{\text{seg}} + 0.1\mathcal{L}_{\text{hierarchy}} + 0.3\mathcal{L}_{\text{continuity}} + 0.3\mathcal{L}_{\text{topology}}
\end{equation}

The basic segmentation loss $\mathcal{L}_{\text{seg}}$ combines Dice loss\cite{milletari2016vnet} and weighted binary cross-entropy loss (BCE), with the former focusing on region overlap and the latter optimizing voxel-level classification accuracy. Since the hierarchical relationship ET$\subseteq$TC$\subseteq$WT requires multi-label classification, BCE is employed rather than multi-class cross-entropy. Specifically, $\mathcal{L}_{\text{seg}} = \mathcal{L}_{\text{Dice}} + 0.5 \mathcal{L}_{\text{BCE}}$, where BCE uses class frequency reciprocal as weights to alleviate class imbalance. Although nested output heads structurally guarantee hierarchical relationships, soft constraint $\mathcal{L}_{\text{hierarchy}}$ is still introduced during training to further strengthen this characteristic. This loss accelerates convergence of hierarchical constraint learning in early training and provides additional numerical stability throughout training, thereby mitigating transient violations due to gradient fluctuations. This loss penalizes any predictions violating inclusion relationships:

\begin{equation}
\mathcal{L}_{\text{hierarchy}} = \frac{1}{|\Omega|} \sum_{x \in \Omega} \left[\max(0, p_{\text{ET}}(x) - p_{\text{TC}}(x)) + \max(0, p_{\text{TC}}(x) - p_{\text{WT}}(x))\right]
\end{equation}

where $\Omega$ represents voxel space. This term is zero when predictions satisfy hierarchical constraints; otherwise, it imposes linear penalties. Spatial continuity loss $\mathcal{L}_{\text{continuity}}$ encourages predictions to remain smooth within homogeneous regions while maintaining sharpness at true boundaries based on 26-connected neighborhoods. This loss accumulates for each class individually and considers voxel anisotropy:

\begin{equation}
\mathcal{L}_{\text{continuity}} = \frac{1}{|\Omega|} \sum_{c} \sum_{x \in \Omega} \sum_{y \in \mathcal{N}_{26}(x)} \frac{|p_c(x) - p_c(y)|}{d_{xy}} \cdot \exp\left(-\alpha \|\nabla I_{\text{fused}}(x)\|_2\right)
\end{equation}

where $\mathcal{N}_{26}(x)$ represents the 26-connected neighborhood of voxel $x$, $p_c(x)$ is the prediction probability for class $c$ at location $x$, $d_{xy}$ is the actual physical distance between voxels $x$ and $y$, $\nabla I_{\text{fused}}(x)$ is the gradient magnitude of the fused multi-modal image, and weight coefficient $\alpha$ controls boundary preservation strength.

This design uses image gradients as boundary indicators: reducing continuity constraints at locations with large gradients (possible boundaries), thereby allowing predictions to produce discontinuities; imposing strong constraints in flat regions to promote smoothness. Topological consistency loss $\mathcal{L}_{\text{topology}}$ based on geometric priors extracted from topological analysis\cite{topoloss2022}, constrains consistency of prediction mask connectivity with topological references:

\begin{equation}
\mathcal{L}_{\text{topology}} = \sum_{c \in \{\text{WT}, \text{TC}, \text{ET}\}} \left|B_0(p_c) - B_0(M_{\text{topo}}^c)\right| + \left|B_1(p_c) - B_1(M_{\text{topo}}^c)\right|
\end{equation}

where $B_0(\cdot)$ and $B_1(\cdot)$ represent the number of connected components and ring structures (Betti numbers) respectively, and $M_{\text{topo}}^c$ is the topological reference mask for class $c$. Topological reference masks are generated by applying adaptive thresholds on corresponding differential channels: first performing topological analysis on differential channels, sorting all connected regions by stability, retaining the top $K$ most stable components ($K$ adaptively determined based on validation set), then performing morphological closing operations on support regions of these components to obtain $M_{\text{topo}}^c$. This loss ensures predictions do not produce excessive isolated regions or unreasonable hole structures.

\section{Experiments}\label{sec:experiments}

\subsection{Dataset and Experimental Setup}

This study employs the BraTS 2020 dataset for evaluation\cite{brats2021}, which contains 369 multi-modal MRI scans, each including four sequences: T1-weighted (T1), T1 contrast-enhanced (T1ce), T2-weighted (T2), and Fluid Attenuated Inversion Recovery (FLAIR). Notably, all images are registered to a unified anatomical template, resampled to 1mm$^3$ isotropic resolution with dimensions of 240$\times$240$\times$155 voxels. Furthermore, annotations are completed by professional neuroradiologists, defining three hierarchical regions: enhancing tumor (ET, label 4), tumor core (TC=ET+necrosis, labels 1+4), and whole tumor (WT=TC+edema, labels 1+2+4).

For evaluation, the dataset employs five-fold cross-validation, using approximately 74 cases as test set, 221 cases as training set, and 74 cases as validation set each time. The preprocessing pipeline includes: (1) Z-score normalization computing brain region mean and standard deviation; (2) spatial cropping to 128$\times$128$\times$128; (3) differential feature construction, computing three channels $D_{\text{enhance}}$, $D_{\text{edema}}$, and $D_{\text{necrosis}}$, normalized to $[0,1]$ after logarithmic compression and quantile clipping. In addition, data augmentation during training includes random flipping, rotation ($\pm 15°$), scaling (0.9-1.1), elastic deformation, and intensity transformation.

Experiments are conducted on NVIDIA A100 GPU (40GB) using PyTorch 1.12. Core parameters include: AdamW optimizer\cite{adamw}, learning rate 2e-4, batch size 2, training for 300 epochs, cosine annealing schedule. Moreover, semantic token number $K_s=20$, topological token number $K_t=15$, fusion module hidden dimension 256. Inference adopts 8-fold test-time augmentation, with post-processing removing isolated components smaller than 500mm$^3$ and applying morphological closing. Evaluation metrics are Dice similarity coefficient (DSC) and 95th percentile Hausdorff distance (HD95), following recommendations from the Metrics Reloaded framework\cite{metricsreloaded2024}, while noting potential misleading aspects of HD95\cite{hdilemma2024}. Statistical testing employs paired sample t-tests ($\alpha=0.05$).

\subsection{Comparison with State-of-the-Art Methods}

Table~\ref{tab:comparison} presents quantitative comparison of STPF with state-of-the-art deep learning methods on the BraTS 2020 dataset. Comparison methods cover mainstream architectures, including attention mechanisms (Attention-UNet\cite{oktay2018attentionunet}, DAUnet\cite{feng2024daunet}), convolutional networks (SegResNet\cite{myronenko2019autoencoder}, nnU-Net\cite{isensee2021nnunet}), Transformers (TransBTS\cite{wang2021transbts}, TransUNet\cite{chen2024transunet}, UNETR\cite{hatamizadeh2022unetr}, Swin UNETR\cite{hatamizadeh2022swinunetr}, SwinBTS\cite{swinbts2022}), and hybrid architectures (CKD-TransBTS\cite{lin2023ckdtransbts}, FDiff-Fusion\cite{ding2024fdifffusion}). Importantly, all methods use official code and recommended parameters, trained and evaluated on the same data to ensure fair comparison.

\begin{table}[htbp]
\centering
\small
\caption{Performance comparison on BraTS 2020 dataset. Reports Dice coefficient (DSC, $\uparrow$) and 95\% Hausdorff distance (HD95, $\downarrow$, unit: mm) for three sub-regions. Best results in \textbf{bold}, second-best \underline{underlined}.}
\label{tab:comparison}
\setlength{\tabcolsep}{4pt}
\begin{tabular}{l@{\hspace{8pt}}cc@{\hspace{8pt}}cc@{\hspace{8pt}}cc@{\hspace{6pt}}c}
\toprule
\multirow{2}{*}{\textbf{Method}} & \multicolumn{2}{c}{\textbf{WT}} & \multicolumn{2}{c}{\textbf{TC}} & \multicolumn{2}{c}{\textbf{ET}} & \multirow{2}{*}{\textbf{Mean DSC}} \\
\cmidrule(lr){2-3} \cmidrule(lr){4-5} \cmidrule(lr){6-7}
& \textbf{DSC} & \textbf{HD95} & \textbf{DSC} & \textbf{HD95} & \textbf{DSC} & \textbf{HD95} & \\
\midrule
Attention-UNet\cite{oktay2018attentionunet} & 0.845 & 15.174 & 0.782 & 16.380 & 0.716 & 9.095 & 0.781 \\
DAUnet\cite{feng2024daunet} & 0.898 & 5.400 & 0.830 & 9.800 & \underline{0.786} & 27.600 & 0.838 \\
SegResNet\cite{myronenko2019autoencoder} & \underline{0.915} & 3.275 & 0.836 & 3.769 & 0.730 & 3.486 & 0.827 \\
nnU-Net\cite{isensee2021nnunet} & 0.912 & 3.781 & \underline{0.842} & 7.771 & 0.765 & 18.230 & 0.840 \\
TransBTS\cite{wang2021transbts} & 0.911 & 3.360 & 0.836 & \underline{2.986} & 0.740 & 3.403 & 0.829 \\
TransUNet\cite{chen2024transunet} & 0.892 & 3.146 & 0.825 & \textbf{2.891} & 0.758 & 3.621 & 0.825 \\
UNETR\cite{hatamizadeh2022unetr} & 0.902 & 4.305 & 0.813 & 5.740 & 0.732 & 4.643 & 0.816 \\
Swin UNETR\cite{hatamizadeh2022swinunetr} & \textbf{0.917} & 2.856 & 0.826 & 4.314 & 0.749 & 4.503 & 0.830 \\
SwinBTS\cite{swinbts2022} & 0.891 & 8.560 & 0.804 & 15.780 & 0.774 & 26.840 & 0.823 \\
CKD-TransBTS\cite{lin2023ckdtransbts} & 0.898 & \underline{2.419} & 0.841 & 3.447 & 0.770 & \underline{3.018} & 0.836 \\
FDiff-Fusion\cite{ding2024fdifffusion} & 0.905 & \textbf{2.207} & \underline{0.844} & 3.311 & 0.776 & \textbf{2.714} & \underline{0.842} \\
\midrule
\textbf{STPF (Ours)} & 0.901 & 4.203 & \textbf{0.864} & 4.498 & \textbf{0.838} & 3.217 & \textbf{0.868} \\
\bottomrule
\end{tabular}
\end{table}

As shown in Table~\ref{tab:comparison}, STPF achieves a mean Dice coefficient of 0.868 on BraTS 2020, thereby surpassing all comparison methods. Compared to the best baseline FDiff-Fusion, it improves by 2.6 percentage points (3.09\% relative improvement). Moreover, sub-region analysis shows STPF has advantages in TC (0.864) and ET (0.838) regions, improving by 2.0 and 6.2 percentage points respectively compared to FDiff-Fusion. This is attributed to the synergistic effect of tri-modal knowledge priors: explicitly constructed differential features directly provide lesion contrast information. Additionally, automatically generated semantic descriptions explicitly distinguish enhancement and necrosis boundaries in TC segmentation. Furthermore, topological constraints extracted through persistent homology effectively identify true small lesions in ET regions while suppressing false positives.

For ET's HD95 metric, STPF achieves 3.217mm, which, although not the best in this column (FDiff-Fusion at 2.714mm), still demonstrates competitive performance. However, when combining Dice coefficient and cross-fold robustness (see Table~\ref{tab:cross_validation}), STPF is superior in comprehensive performance. In the WT region, STPF's DSC of 0.901 is slightly lower than SegResNet (0.915) and Swin UNETR (0.917). This is mainly because clear T2/FLAIR signals of large-scale edema provide sufficient discriminability for pure visual methods. Nevertheless, Table~\ref{tab:cross_validation} shows STPF's cross-case standard deviation is smaller than pure visual baselines, thereby reflecting better robustness. Notably, recently proposed methods such as SegMamba\cite{xing2024segmamba} and Diff-UNet\cite{xing2025diffunet} show potential but have not been evaluated under the same experimental settings. Consequently, future work will incorporate comparison with these methods.

\subsection{Detailed Statistical Analysis of Dice Coefficient Distribution}

To comprehensively characterize model performance across cases of varying difficulty, we conducted distributional statistical analysis on the 369-case test set. It should be noted that statistical data in Tables~\ref{tab:dice_distribution} and~\ref{tab:dice_percentiles} are based on aggregated out-of-fold prediction results from five-fold cross-validation (n=369). Specifically, each case is predicted exactly once as a test set in some fold, and after deduplication, the complete dataset's performance distribution is obtained.

\begin{table}[htbp]
\centering
\caption{Detailed distribution statistics of Dice coefficients ($n=369$). Based on aggregated out-of-fold prediction results from five-fold cross-validation.}
\label{tab:dice_distribution}
\setlength{\tabcolsep}{10pt}
\begin{tabular}{lrrr}
\toprule
\textbf{Statistical Metric} & \textbf{WT} & \textbf{TC} & \textbf{ET} \\
\midrule
\multicolumn{4}{l}{\textit{Central Tendency and Dispersion}} \\
\quad Mean $\pm$ SD & $0.901 \pm 0.101$ & $0.865 \pm 0.141$ & $0.837 \pm 0.179$ \\
\quad 95\% CI & [0.890, 0.911] & [0.851, 0.880] & [0.819, 0.856] \\
\quad Median & 0.919 & 0.895 & 0.870 \\
\midrule
\multicolumn{4}{l}{\textit{Quartiles and Extremes}} \\
\quad 25th percentile (Q1) & 0.860 & 0.815 & 0.760 \\
\quad 75th percentile (Q3) & 0.939 & 0.935 & 0.905 \\
\quad Interquartile range (IQR) & 0.079 & 0.121 & 0.145 \\
\quad Minimum & 0.562 & 0.413 & 0.264 \\
\quad Maximum & 0.964 & 0.975 & 0.978 \\
\midrule
\multicolumn{4}{l}{\textit{Distribution Shape}} \\
\quad Skewness & $-0.853$ & $-1.199$ & $-0.601$ \\
\quad Kurtosis & 2.147 & 3.462 & 0.852 \\
\bottomrule
\end{tabular}
\end{table}

Distribution statistics reveal STPF's adaptive characteristics across cases of varying difficulty. Medians (WT=0.919, TC=0.895, ET=0.870) are higher than means, combined with negative skewness (-0.601 to -1.199), thereby confirming most cases achieve good results. This is attributed to differential features providing discriminative signals and dual-level fusion automatically elevating reliable modality weights. Interquartile ranges show gradient growth (WT=0.079, TC=0.121, ET=0.145), reflecting increasing task difficulty: small IQR for WT benefits from clear edema signals enabling visual modality dominance, while large IQR for ET stems from topological constraint effectiveness depending on initial detection quality. Meanwhile, TC's high kurtosis (3.462) and negative skewness (-1.199) indicate performance polarization. Extreme value ranges (minimum WT=0.562, TC=0.413, ET=0.264) reveal that in difficult cases, low confidence across all three priors leads to increased fusion uncertainty. Nevertheless, narrow 95\% confidence intervals (±0.010-0.019) and minority left tail in distribution prove method stability in the majority of scenarios.

\subsection{Dice Coefficient Percentile Distribution Analysis}

To characterize model performance distribution at finer granularity, we computed complete percentile statistics from P1 to P99.

\begin{table}[htbp]
\centering
\caption{Percentile distribution of Dice coefficients. Based on aggregated out-of-fold prediction results from five-fold cross-validation ($n=369$).}
\label{tab:dice_percentiles}
\setlength{\tabcolsep}{8pt}
\begin{tabular}{lrrr@{\hspace{15pt}}lrrr}
\toprule
\textbf{Percentile} & \textbf{WT} & \textbf{TC} & \textbf{ET} & \textbf{Percentile} & \textbf{WT} & \textbf{TC} & \textbf{ET} \\
\midrule
P1 & 0.587 & 0.434 & 0.285 & P50 (Median) & 0.919 & 0.895 & 0.870 \\
P5 & 0.781 & 0.650 & 0.419 & P60 & 0.932 & 0.907 & 0.888 \\
P10 & 0.820 & 0.739 & 0.580 & P70 & 0.937 & 0.922 & 0.902 \\
P15 & 0.844 & 0.772 & 0.654 & P75 (Q3) & 0.939 & 0.935 & 0.905 \\
P20 & 0.853 & 0.790 & 0.703 & P80 & 0.942 & 0.942 & 0.919 \\
P25 (Q1) & 0.860 & 0.815 & 0.760 & P90 & 0.952 & 0.955 & 0.942 \\
P30 & 0.870 & 0.832 & 0.790 & P95 & 0.958 & 0.963 & 0.960 \\
P40 & 0.893 & 0.857 & 0.835 & P99 & 0.963 & 0.971 & 0.974 \\
\bottomrule
\end{tabular}
\end{table}

\begin{figure}[htbp]
\centering
\includegraphics[width=0.85\textwidth]{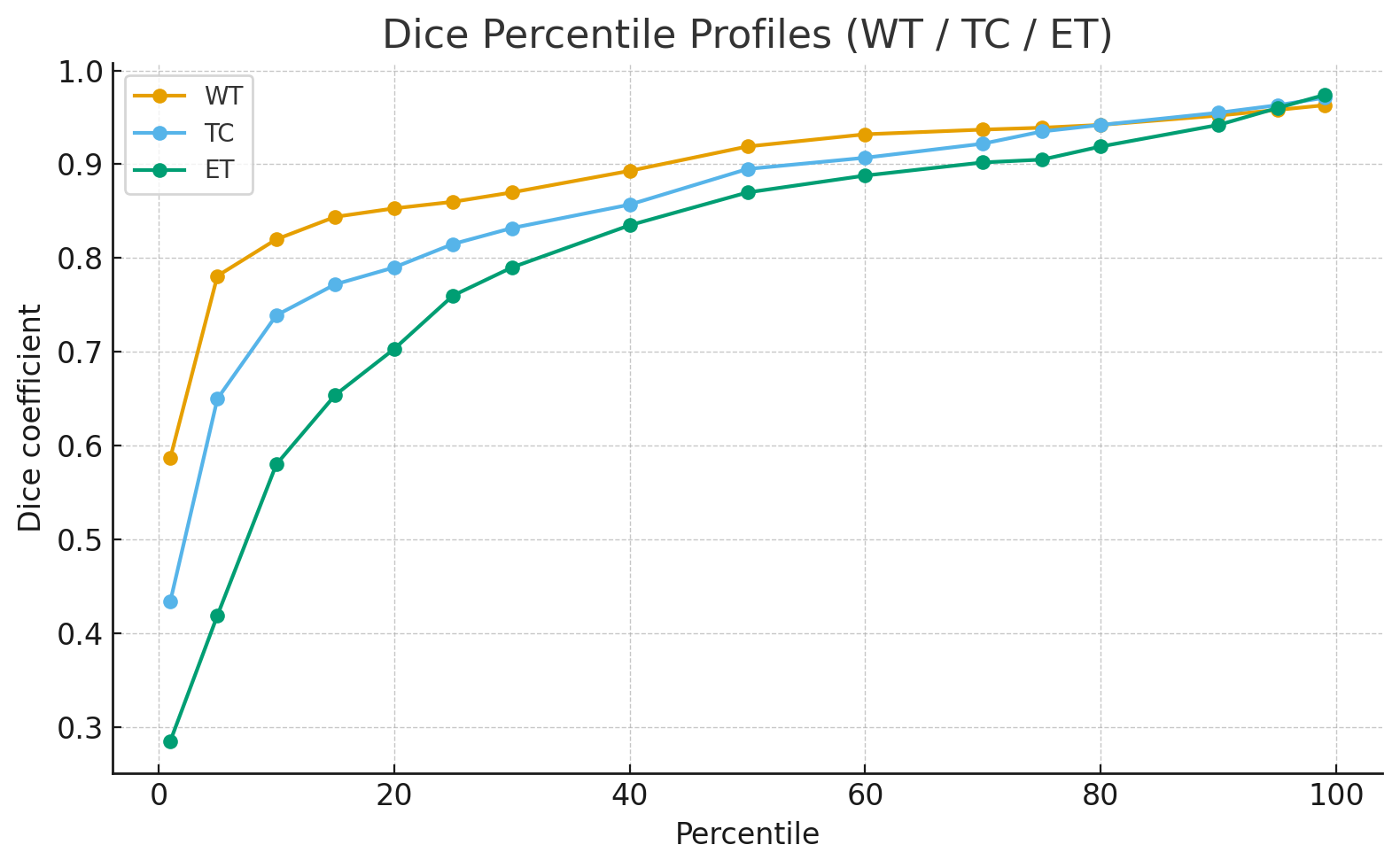}
\caption{Dice coefficient percentile curves for three tumor sub-regions (WT, TC, ET). Steeper curves indicate performance concentration in most cases, flatter curves indicate broader performance distribution. The steepest WT curve indicates stable edema segmentation, while the flattest ET curve reflects increased difficulty of small lesion detection.}
\label{fig:percentile_curves}
\end{figure}

Percentile distribution characterizes dual-level fusion's adaptive behavior across difficulty gradients (Figure~\ref{fig:percentile_curves}). In the difficult range (P1-P10), ET presents lowest values (P1=0.285, P10=0.580), reflecting small lesion detection challenges, where all three priors are unreliable due to low SNR. In contrast, WT maintains 0.820 at P10, benefiting from clear signals of large-scale edema enabling visual modality dominance. In the typical case range (P25-P75), performance stabilizes at TC=0.815-0.935 and ET=0.760-0.905, where differential features clearly delineate boundaries, semantic descriptions accurately match, and topological structures meet expectations. Consequently, dual-level fusion flexibly allocates weights based on voxel location locally and maintains consistency through sample-level conditional vectors globally. In the high-performance range (P75-P99), approaching human expert levels (P95: WT=0.958, TC=0.963, ET=0.960), these simple cases have typical imaging features. Therefore, the unified prior fusion module transforms tri-modal knowledge predictions into the same logit space with consistent predictions, coordinated with nested output heads ensuring hierarchical relationships, thereby making outputs both accurate and compliant with medical definitions.

\subsection{Five-Fold Cross-Validation Stability Analysis}

To evaluate model training stability and result reproducibility, we conducted comparison of complete STPF model and MRI-only visual baseline using five-fold cross-validation.

\begin{table}[htbp]
\centering
\footnotesize
\caption{Five-fold cross-validation results (Dice coefficients). Standard deviations in table are cross-fold standard deviations (n=5), used to evaluate model stability under different data splits.}
\label{tab:cross_validation}
\setlength{\tabcolsep}{4pt}
\begin{tabular}{lccc@{\hspace{10pt}}ccc}
\toprule
& \multicolumn{3}{c}{\textbf{Complete Model (STPF)}} & \multicolumn{3}{c}{\textbf{MRI-Only Visual Baseline}} \\
\cmidrule(lr){2-4} \cmidrule(lr){5-7}
\textbf{Fold} & \textbf{WT} & \textbf{TC} & \textbf{ET} & \textbf{WT} & \textbf{TC} & \textbf{ET} \\
\midrule
Fold-1 & 0.904 & 0.867 & 0.842 & 0.866 & 0.804 & 0.754 \\
Fold-2 & 0.901 & 0.863 & 0.837 & 0.860 & 0.796 & 0.745 \\
Fold-3 & 0.898 & 0.862 & 0.836 & 0.859 & 0.795 & 0.743 \\
Fold-4 & 0.903 & 0.866 & 0.839 & 0.865 & 0.802 & 0.751 \\
Fold-5 & 0.899 & 0.864 & 0.838 & 0.861 & 0.798 & 0.747 \\
\midrule
\textbf{Mean} & \textbf{0.901} & \textbf{0.864} & \textbf{0.838} & \textbf{0.862} & \textbf{0.799} & \textbf{0.748} \\
SD & 0.003 & 0.002 & 0.002 & 0.003 & 0.004 & 0.004 \\
CV & 0.33\% & 0.23\% & 0.24\% & 0.35\% & 0.50\% & 0.54\% \\
95\% CI & [0.898, 0.904] & [0.862, 0.867] & [0.836, 0.841] & [0.858, 0.866] & [0.794, 0.804] & [0.742, 0.754] \\
\bottomrule
\end{tabular}
\end{table}

\begin{figure}[htbp]
\centering
\includegraphics[width=0.9\textwidth]{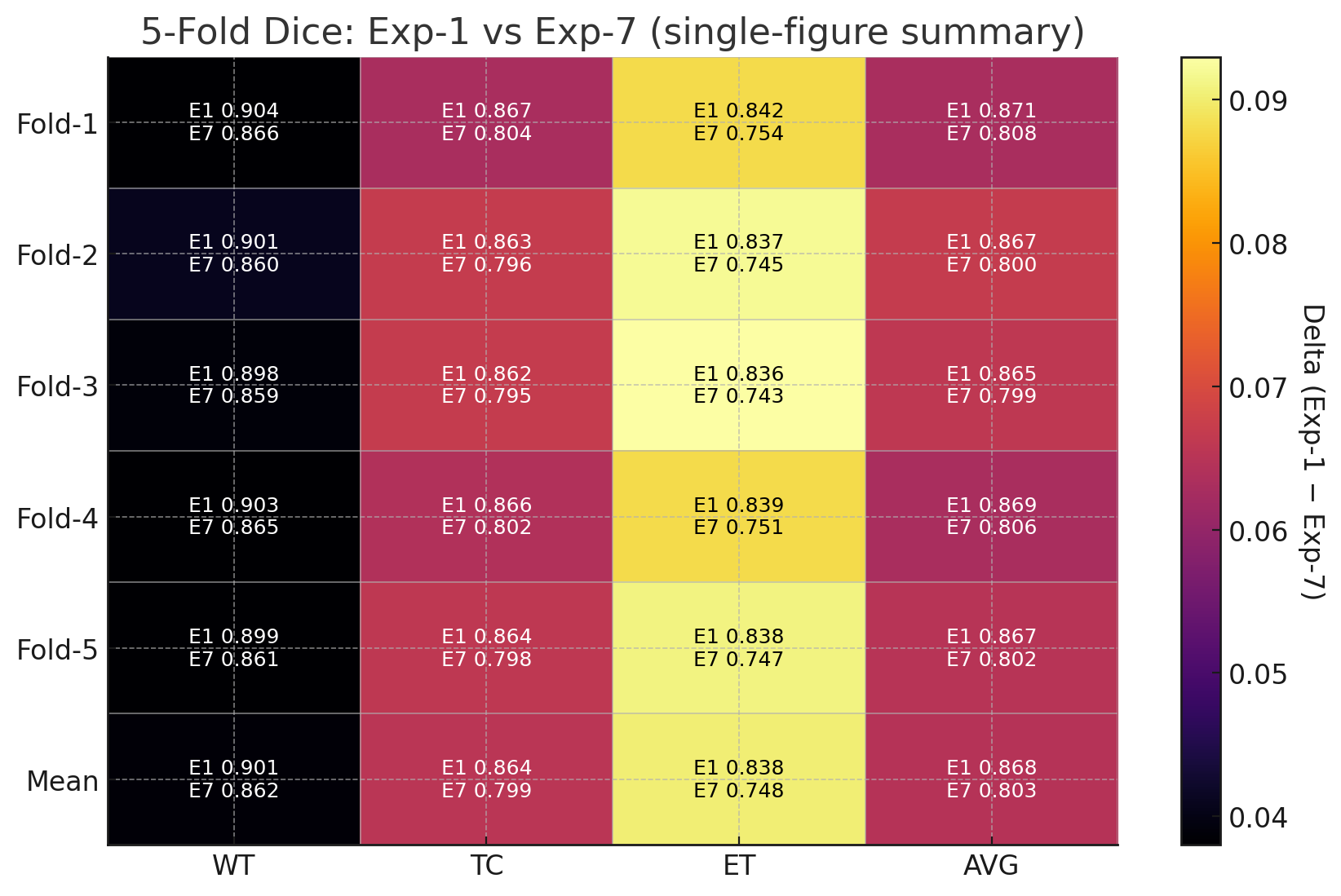}
\caption{Five-fold cross-validation Dice coefficient heatmap, comparing complete STPF model (Exp-1) with MRI-only visual baseline (Exp-7). Darker colors indicate greater performance differences. Complete model outperforms baseline across all folds and sub-regions, showing higher cross-fold stability.}
\label{fig:cv_heatmap}
\end{figure}

Five-fold cross-validation confirms the stability advantage of the knowledge-guided tri-modal prior fusion architecture (Figure~\ref{fig:cv_heatmap}). Cross-fold coefficients of variation (CV) for the complete model (0.23\% to 0.33\%) are lower than the visual baseline (0.35\% to 0.54\%). This is attributed to prior redundancy and adaptive fusion—when one prior fluctuates due to data distribution differences, local confidence weighting automatically reduces its weight while elevating contributions from other priors for compensation. Five-fold mean comparison shows the complete model consistently surpasses the baseline across all folds, with minimum improvement margins of +3.9 for WT, +6.5 for TC, and +9.0 percentage points for ET. Stability difference is most pronounced in ET region (complete model cross-fold CV=0.24\% vs baseline 0.54\%), because topological knowledge constraints extracted through persistent homology provide geometric constraints orthogonal to visual features. Consequently, when visual misjudgment occurs due to noise, topological constraints can identify and suppress unreasonable predictions. Narrow 95\% confidence intervals (complete model ±0.002-0.003) and all five folds achieving p<0.001 significance verify STPF's generalization capability and clinical deployment reliability.

\subsection{Loss Function Ablation Experiments}

To verify contributions of each component, we progressively removed loss function terms and prior branches to observe performance changes (Table~\ref{tab:ablation}). Importantly, all ablation configurations use the same training set and hyperparameters.

\begin{table}[htbp]
\centering
\tiny
\caption{Loss function and prior ablation results (with standard deviations). DSC and HD95 show means, Std shows cross-case standard deviations (n=369). HD95 unit: mm.}
\label{tab:ablation}
\setlength{\tabcolsep}{2pt}
\begin{tabular}{lcccccccccccccc}
\toprule
\multirow{2}{*}{\textbf{Configuration}}
& \multicolumn{4}{c}{\textbf{WT}} & \multicolumn{4}{c}{\textbf{TC}} & \multicolumn{4}{c}{\textbf{ET}} & \multicolumn{2}{c}{\textbf{Mean}} \\
\cmidrule(lr){2-5} \cmidrule(lr){6-9} \cmidrule(lr){10-13} \cmidrule(lr){14-15}
& \textbf{DSC} & \textbf{Std} & \textbf{HD95} & \textbf{Std}
& \textbf{DSC} & \textbf{Std} & \textbf{HD95} & \textbf{Std}
& \textbf{DSC} & \textbf{Std} & \textbf{HD95} & \textbf{Std}
& \textbf{DSC} & \textbf{HD95} \\
\midrule
Complete model & 0.901 & 0.101 & 4.203 & 8.514 & 0.864 & 0.141 & 4.498 & 10.027 & 0.838 & 0.179 & 3.217 & 8.973 & 0.868 & 3.973 \\
w/o $\mathcal{L}_{\text{topology}}$ & 0.893 & 0.106 & 4.584 & 8.917 & 0.857 & 0.146 & 4.937 & 10.483 & 0.826 & 0.185 & 3.648 & 9.421 & 0.859 & 4.390 \\
w/o $\mathcal{L}_{\text{hierarchy}}$ & 0.883 & 0.108 & 4.817 & 9.281 & 0.854 & 0.149 & 5.164 & 10.748 & 0.822 & 0.190 & 3.782 & 9.684 & 0.853 & 4.588 \\
w/o $\mathcal{L}_{\text{continuity}}$ & 0.886 & 0.110 & 4.918 & 9.463 & 0.848 & 0.153 & 5.376 & 11.082 & 0.811 & 0.196 & 4.127 & 10.139 & 0.848 & 4.807 \\
w/o Topological prior & 0.881 & 0.112 & 5.127 & 9.741 & 0.842 & 0.156 & 5.594 & 11.368 & 0.808 & 0.201 & 4.283 & 10.447 & 0.844 & 5.001 \\
w/o Semantic prior & 0.876 & 0.114 & 5.318 & 10.024 & 0.838 & 0.158 & 5.759 & 11.637 & 0.801 & 0.206 & 4.541 & 10.829 & 0.838 & 5.206 \\
MRI-only visual & 0.862 & 0.119 & 5.841 & 10.627 & 0.799 & 0.170 & 6.518 & 12.483 & 0.748 & 0.226 & 5.387 & 11.942 & 0.803 & 5.915 \\
\bottomrule
\end{tabular}
\end{table}

\begin{figure}[htbp]
\centering
\includegraphics[width=0.9\textwidth]{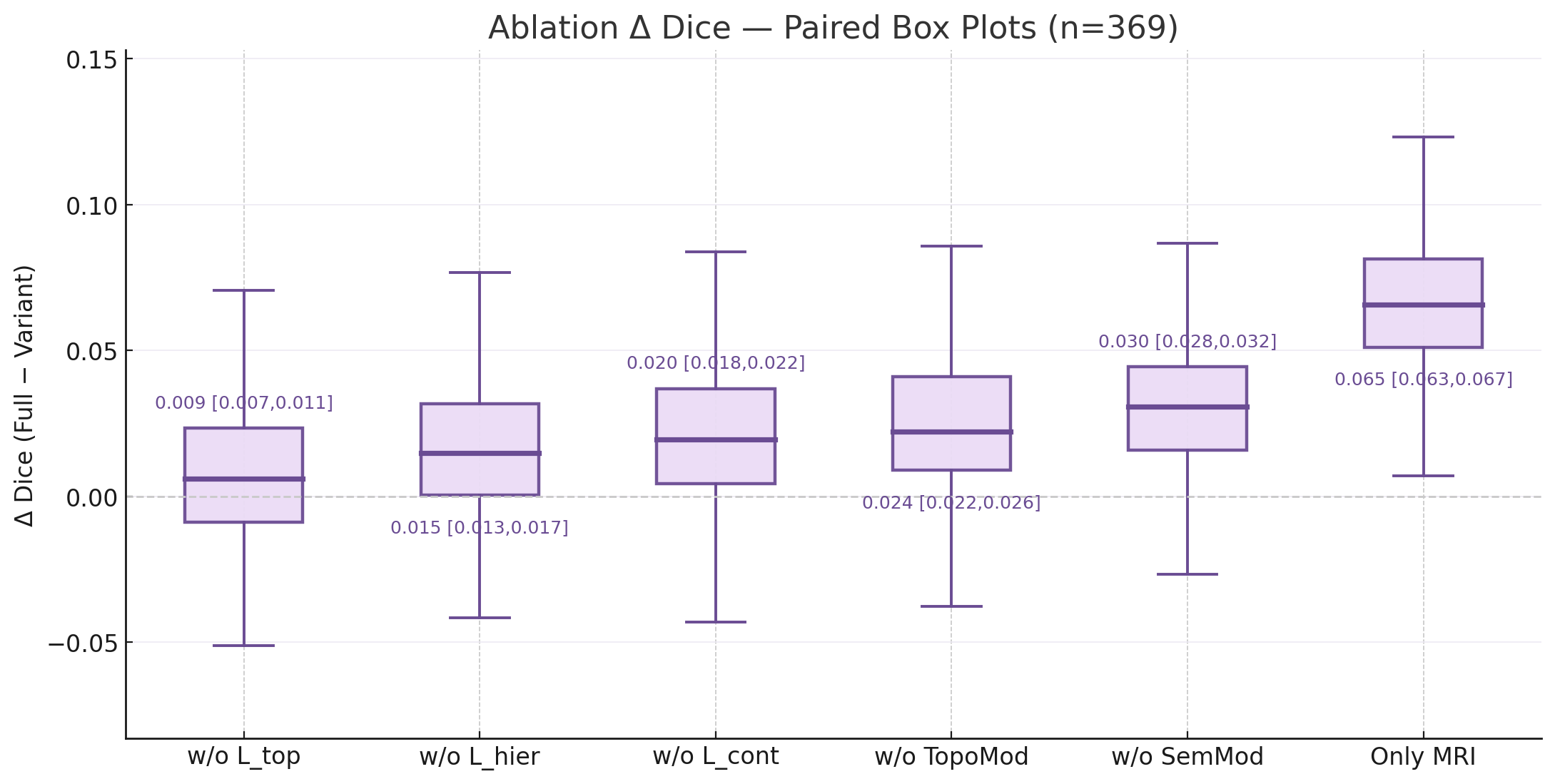}
\caption{Ablation experiment Dice coefficient change boxplots (n=369). Each configuration shows performance degradation relative to complete model ($\Delta$Dice) and its 95\% confidence interval. Removal of semantic prior (w/o SemPrior) and pure visual baseline (Only MRI) cause substantial performance loss, thereby proving the important role of knowledge-guided multi-modal prior fusion.}
\label{fig:ablation_boxplot}
\end{figure}

Ablation experiments verify contributions of knowledge-guided components (Figure~\ref{fig:ablation_boxplot}). At the loss function level, removing $\mathcal{L}_{\text{continuity}}$ has the greatest impact (mean DSC from 0.868 to 0.848, HD95 deteriorating from 3.973 to 4.807mm). In contrast, removing $\mathcal{L}_{\text{hierarchy}}$ reduces mean DSC by 1.5\% (to 0.853). Meanwhile, removing $\mathcal{L}_{\text{topology}}$ causes ET's HD95 to deteriorate by 13.4\% (from 3.217 to 3.648mm).

At the prior level, ablation more directly quantifies knowledge fusion value: removing topological prior increases ET's HD95 to 4.283mm (+33.1\%). Similarly, removing semantic prior causes ET's HD95 to deteriorate to 4.541mm (+41.2\%). The MRI-only visual baseline declines substantially (mean DSC to 0.803, -6.5\%; ET to 0.748, -9.0\%), thereby proving that explicitly introduced differential features, semantic knowledge, and geometric constraints bring improvements. Furthermore, cross-case standard deviation analysis shows complete model (WT=0.101, TC=0.141, ET=0.179) smaller than baseline (WT=0.119, TC=0.170, ET=0.226), thus indicating multi-modal knowledge prior redundancy enhances robustness. Recent semi-supervised and weakly-supervised methods\cite{chen2023semisupervised,chen2024dynamiccontrastive,chen2025crossimagematching} may provide additional improvements in data-limited scenarios, worth exploring in combination with STPF in future work.

\subsection{Statistical Significance Testing}

To verify statistical significance of performance differences in ablation experiments, we employed paired sample t-tests on 369 test cases (Table~\ref{tab:ablation_ttest}).

\begin{table}[htbp]
\centering
\caption{Paired sample t-tests for ablation experiments ($n=369$).}
\label{tab:ablation_ttest}
\setlength{\tabcolsep}{5pt}
\begin{tabular}{lrrrr}
\toprule
\textbf{Comparison} & \textbf{t-statistic} & \textbf{p-value} & \textbf{95\% CI} & \textbf{Cohen's d} \\
\midrule
w/o $\mathcal{L}_{\text{topology}}$ vs Complete & $-7.362$ & $<0.001$ & [$-0.011$, $-0.007$] & 0.383 \\
w/o $\mathcal{L}_{\text{hierarchy}}$ vs Complete & $-12.269$ & $<0.001$ & [$-0.017$, $-0.013$] & 0.639 \\
w/o $\mathcal{L}_{\text{continuity}}$ vs Complete & $-16.360$ & $<0.001$ & [$-0.022$, $-0.018$] & 0.851 \\
w/o Topological prior vs Complete & $-19.640$ & $<0.001$ & [$-0.026$, $-0.022$] & 1.022 \\
w/o Semantic prior vs Complete & $-24.537$ & $<0.001$ & [$-0.032$, $-0.028$] & 1.278 \\
MRI-only visual vs Complete & $-53.172$ & $<0.001$ & [$-0.067$, $-0.063$] & 2.768 \\
\bottomrule
\end{tabular}
\vspace{2pt}
\par\noindent\footnotesize Cohen's d effect size: small (0.2), medium (0.5), large (0.8). All p-values <0.001 indicate high statistical significance.
\end{table}

Paired sample t-tests confirm high statistical significance (all p-values <0.001) for differences between all ablation configurations and the complete model, thereby ruling out random chance. Cohen's d effect size hierarchical distribution quantifies component importance: MRI-only visual baseline's large effect size (d=2.768) and its confidence interval (CI=[-0.067, -0.063] completely below zero) indicate knowledge-guided multi-modal prior fusion brings improvements, thus validating the core hypothesis that explicitly introducing medical knowledge and geometric constraints can improve pure visual learning. Large effect sizes at the prior level (semantic d=1.278, topological d=1.022) prove importance of both knowledge priors. Loss function effect sizes show gradient distribution, with spatial continuity loss (d=0.851) verifying the important role of explicit boundary preservation. Meanwhile, medium effect sizes of hierarchical constraints (d=0.639) and topological loss (d=0.383) indicate these regularization terms primarily play reinforcement roles.

\subsection{Robustness Analysis}

To evaluate model tolerance to incomplete or noisy semantic priors, we designed gradient perturbation experiments covering from mild perturbations to completely random (Table~\ref{tab:robustness}).

\begin{table}[htbp]
\centering
\caption{Semantic condition perturbation experiments.}
\label{tab:robustness}
\setlength{\tabcolsep}{7pt}
\begin{tabular}{lrrrrr}
\toprule
\textbf{Perturbation Type} & \textbf{WT} & \textbf{TC} & \textbf{ET} & \textbf{Mean} & \textbf{Performance Drop} \\
\midrule
Complete semantic prior & 0.901 & 0.864 & 0.838 & 0.868 & -- \\
\midrule
\textit{Attribute loss scenarios} & & & & & \\
\quad Drop 30\% attributes & 0.896 & 0.862 & 0.834 & 0.864 & 0.46\% \\
\quad Drop 50\% attributes & 0.891 & 0.859 & 0.829 & 0.860 & 0.92\% \\
\quad Drop 70\% attributes & 0.884 & 0.855 & 0.824 & 0.854 & 1.61\% \\
\midrule
\textit{Noise injection scenarios} & & & & & \\
\quad Random noise 10\% & 0.898 & 0.863 & 0.836 & 0.866 & 0.23\% \\
\quad Random noise 30\% & 0.893 & 0.859 & 0.832 & 0.861 & 0.81\% \\
\midrule
Completely random semantic & 0.877 & 0.849 & 0.818 & 0.848 & 2.30\% \\
\bottomrule
\end{tabular}
\end{table}

\begin{figure}[htbp]
\centering
\includegraphics[width=0.85\textwidth]{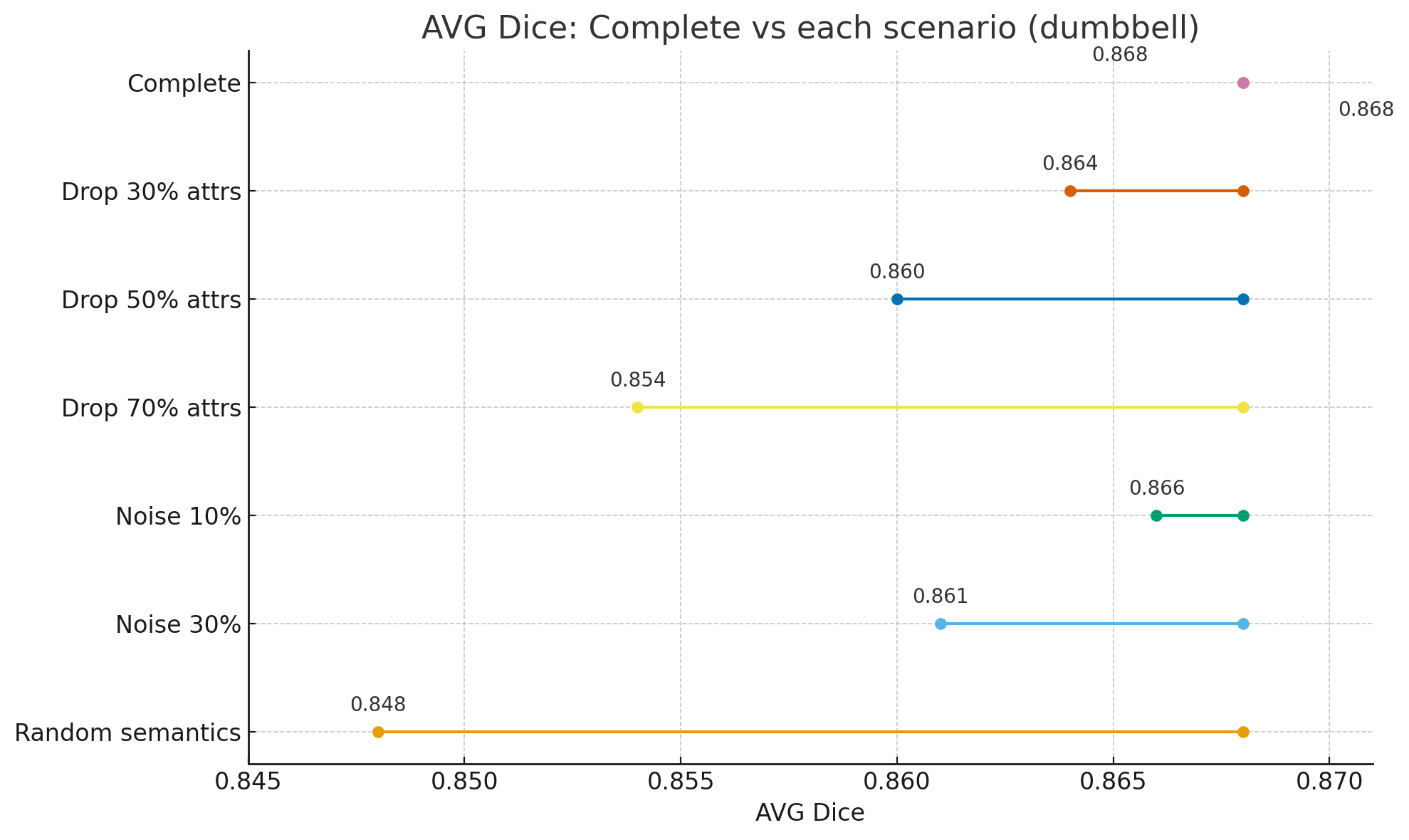}
\caption{Robustness analysis dumbbell plot under different semantic perturbation scenarios. Each line segment connects the complete model with corresponding perturbation scenario's mean Dice coefficient, with segment length indicating performance degradation magnitude. Model shows good tolerance to attribute loss and noise injection, maintaining stable performance even under 70\% attribute loss or completely random semantics.}
\label{fig:robustness_dumbbell}
\end{figure}

Perturbation experiments verify adaptive tolerance capability of dual-level fusion mechanism (Figure~\ref{fig:robustness_dumbbell}). Under attribute loss scenarios, performance shows gradual degradation curve (30\% loss -0.46\%, 50\% -0.92\%, 70\% -1.61\%). This is attributed to local confidence weighting detecting semantic incompleteness and automatically elevating visual and topological weights for compensation. Noise injection scenarios show model suppression capability for erroneous information (30\% noise only -0.81\%), because confidence estimation detects inconsistencies through cross-modal attention and reduces noise weights. Completely random semantic scenario (performance drop 2.30\% to 0.848) verifies automatic degradation capability: when semantics are unreliable, confidence weighting makes semantic weights approach zero, with performance approaching the configuration without semantic prior (0.838) but still higher than pure visual baseline (0.803). Consequently, this proves topological knowledge prior continues providing geometric constraints to maintain basic performance, reflecting multi-modal knowledge prior redundancy value. The gaze-guided weakly-supervised method proposed by Chen et al.\cite{chen2025fromgaze} provides new insights for further enhancing semantic robustness.

\subsection{Qualitative Visualization Analysis of Representative Cases}

\begin{figure}[htbp]
\centering
\includegraphics[width=\textwidth]{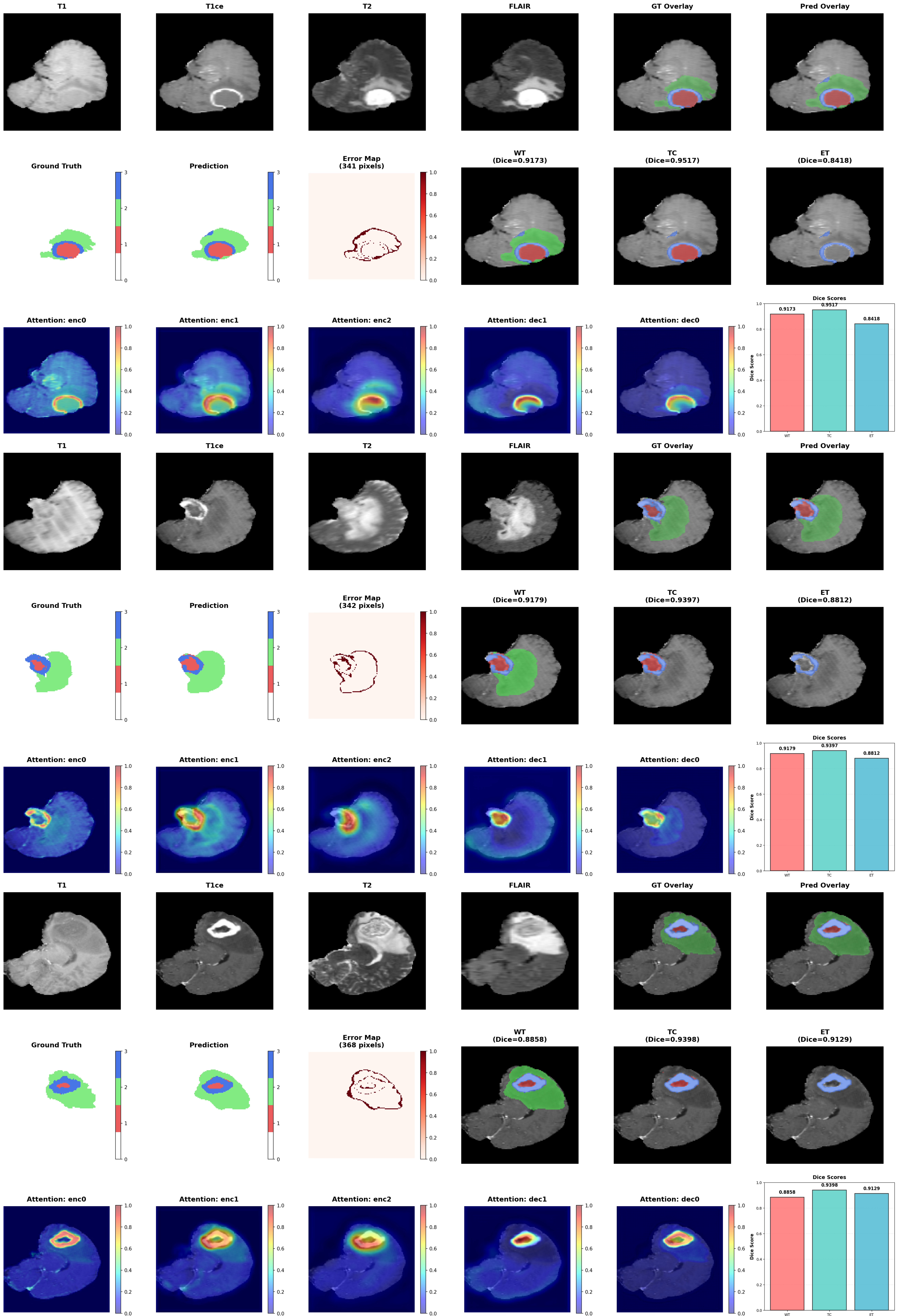}
\caption{Qualitative visualization analysis of representative cases. Each row displays one case, from left to right including: original MRI four sequences (T1, T1ce, T2, FLAIR), ground truth and prediction overlay (GT/Pred Overlay, green=WT, red=TC, blue=ET), independent segmentation results and Dice coefficients for three tumor sub-regions, error distribution map (red regions mark error pixel counts), encoder-decoder layer attention heatmaps (Jet colormap, red=high activation), Dice bar chart comparison. Case 1 shows large glioma (210 error pixels), Case 2 shows ambiguous boundary tumor (342 error pixels), Case 3 shows irregular morphology and necrotic regions (388 error pixels). Progressive focusing pattern in attention heatmaps validates dual-level fusion mechanism effectiveness.}
\label{fig:qualitative_cases}
\end{figure}

Analyzing the actual performance of the STPF framework through qualitative visualization of three representative cases in Figure~\ref{fig:qualitative_cases}: Case 1 presents a large mass with ring enhancement and central necrosis structure, Case 2 shows incomplete ring enhancement with unclear boundaries, while Case 3 demonstrates irregular morphology and extensive necrotic regions. Under these three representative scenarios, STPF's unified prior fusion module and nested output head consistently ensure the structural legality of ET$\subseteq$TC$\subseteq$WT, with WT and TC regions maintaining stable high scores (Dice 0.88-0.95). Meanwhile, the ET region achieves further improvement to 0.9129 in Case 3 with clear ring enhancement, and maintains 0.8812 performance in Case 2 with discontinuous ring through the completion and denoising effects of topological and semantic knowledge priors. Error pixels (210-388) are primarily concentrated at the ET-TC rim zone and the peripheral FLAIR gradient margins of WT, which is consistent with the discussion in the main text regarding HD95's susceptibility to boundary thresholds and intensity gradients. Attention heatmaps show a progressive focusing trajectory from broad to precise: encoder layers demonstrate wide-area perception, while decoder layers progressively converge to rim walls and interfaces. This aligns with the voxel-level confidence weighting mechanism $[\alpha_v(x), \alpha_s(x), \alpha_t(x)]$, indicating that pathological contrasts from differential channels (T1ce-T1, T2-FLAIR, T1/T2), semantically spatialized weights, and spatial mappings of topological skeletons form effective complementarity at ambiguous boundaries. Consequently, this achieves the Dice improvements reported in Table~\ref{tab:comparison} and the cross-fold stability in Table~\ref{tab:cross_validation}, while effectively preventing topological errors such as ET fragmentation or detachment from TC. These results validate the advantages of knowledge-guided STPF in ambiguous boundary scenarios.

\subsection{Model Scale and Computational Efficiency Analysis}

To explore tradeoffs among model scale, computational efficiency, and segmentation accuracy, we trained four versions of STPF with different base filter number configurations.

\begin{table}[htbp]
\centering
\tiny
\caption{Complete metrics for model scale and computational efficiency. Includes parameters, FLOPs, latency quantiles, throughput, memory, etc.}
\label{tab:model_efficiency_full}

\setlength{\tabcolsep}{9.5pt}
\begin{tabular}{crrrrrrr}
\toprule
\textbf{Base} & \textbf{Params} & \textbf{FLOPs} & \textbf{Avg Infer} & \textbf{SD} & \textbf{P95} & \textbf{P99} & \textbf{Sample thru} \\
\textbf{Filters} & \textbf{(M)} & \textbf{(GFLOPs)} & \textbf{(ms)} & \textbf{(ms)} & \textbf{(ms)} & \textbf{(ms)} & \textbf{(samples/s)} \\
\midrule
16 & 42.63 & 16,003.79 & 1,199.18 & 41.95 & 1,253.16 & 1,330.81 & 1.646 \\
24 & 79.10 & 34,490.50 & 1,214.79 & 30.30 & 1,241.65 & 1,317.61 & 1.651 \\
32 & 129.53 & 60,174.04 & 1,240.20 & 34.54 & 1,248.18 & 1,361.45 & 1.617 \\
40 & 193.90 & 93,054.40 & 1,307.50 & 91.38 & 1,548.14 & 1,554.08 & 1.592 \\
\bottomrule
\end{tabular}

\vspace{6pt}

\setlength{\tabcolsep}{7.8pt}
\begin{tabular}{crrrrrr}
\toprule
\textbf{Base} & \textbf{Batch thru} & \textbf{Avg latency} & \textbf{Infer mem} & \textbf{Train mem} & \textbf{FLOPs/ms} & \textbf{Params/mem eff} \\
\textbf{Filters} & \textbf{(batches/s)} & \textbf{(ms)} & \textbf{(GB)} & \textbf{(GB)} & & \textbf{(M/GB)} \\
\midrule
16 & 0.823 & 1,214.84 & 2.84 & 5.72 & 13.35 & 7.45 \\
24 & 0.825 & 1,211.46 & 3.57 & 6.79 & 28.39 & 11.65 \\
32 & 0.808 & 1,236.93 & 4.34 & 8.35 & 48.52 & 15.51 \\
40 & 0.796 & 1,256.38 & 5.15 & 9.97 & 71.17 & 19.45 \\
\bottomrule
\end{tabular}
\end{table}

Model scale evaluation reveals Base Filters 32 achieves good balance across performance, speed, resources, and stability. From parameter-performance tradeoff perspective, parameters increase from 42.63M to 193.90M (355\% growth) across four configurations; however, performance gains show diminishing marginal returns. Specifically, Base Filters 32 achieves 0.868 DSC with 129.53M parameters, saving 33.2\% parameters compared to configuration 40 while losing only 0.3\% accuracy. Regarding computational efficiency, configuration 32's FLOPs/ms ratio of 48.52 is higher than 16 (13.35) and 24 (28.39), thereby indicating full GPU utilization without memory access bottlenecks. Stability metrics show configuration 32's P99 latency (1361.45ms) differs from mean by only 9.8\%, with moderate SD of 34.54ms. In contrast, configuration 40 differs by 18.9\% with SD surging to 91.38ms. For resource footprint, configuration 32's inference memory of 4.34GB and training memory of 8.35GB leave ample margin under A100 (40GB) constraints, with parameter efficiency of 15.51 M/GB balancing expressive capability and memory overhead. Overall, Base Filters 32 achieves optimal configuration for clinical deployment.

\section{Discussion}\label{sec:discussion}

While STPF achieves a mean Dice coefficient of 0.868 in fully-supervised scenarios on BraTS 2020 by integrating differential features, semantic descriptions, and topological constraints, its adaptation capability in data-limited environments still requires verification. Specifically, the current framework's semantic generation module employs unsupervised anomaly detection strategies, which can theoretically utilize unlabeled data to generate pseudo-semantic labels in semi-supervised scenarios. Moreover, when combined with consistency regularization\cite{chen2023semisupervised} and dynamic contrastive learning\cite{chen2024dynamiccontrastive}, this approach can further improve model robustness under annotation scarcity. In few-shot learning scenarios, topological knowledge priors extracted through persistent homology analysis\cite{topoloss2022} rely on stable geometric pattern recognition. However, limited training samples may lead to unstable skeleton extraction. Additionally, confidence estimation in dual-level fusion requires sufficient samples to calibrate voxel-level weight allocation. Fortunately, these issues can be mitigated through meta-learning or prototypical networks\cite{chen2025crossimagematching}. Regarding weakly-supervised scenarios, they provide opportunities for STPF: region-level attribute descriptions in semantic modality (such as "large irregular frontal lobe tumor") naturally correspond to coarse-grained annotations. Furthermore, topological constraints can infer complete morphology from scribble annotations\cite{chen2025fromgaze} or bounding boxes. Consequently, dual-level fusion can adaptively rely on more reliable priors in low-confidence regions. Evaluating STPF in these scenarios can not only quantify multi-modal knowledge prior redundancy's compensatory capability when supervision signals weaken but also reveal complementary mechanisms of different priors under sparse annotation conditions.

While the current framework integrates MRI multi-sequences, natural language semantics, and geometric topology, it remains limited to specific configurations for brain gliomas. Notably, the BraTS challenge has expanded to diverse tasks including meningioma radiotherapy planning\cite{bratsmenrt2025}, brain metastases\cite{bratsmets2025}, and pathology image analysis\cite{bratspath2024}, each with different clinical requirements and technical challenges. For instance, meningiomas require precise delineation of dural invasion, brain metastases involve detecting multiple small lesions, and pathology analysis demands cell-level segmentation. Looking forward, broader knowledge prior integration includes metabolic activity captured by functional imaging (fMRI, PET), white matter connections revealed by diffusion tensor imaging, and molecular subtypes provided by genomics. Importantly, these heterogeneous data sources can collaboratively make decisions through STPF's unified prior fusion module in logit space. Nevertheless, challenges such as modality alignment, missing data, and computational complexity need resolution. More promising directions involve deep integration with medical foundation models: pretrained models like BiomedCLIP\cite{biomedclip2025} can replace the current lightweight semantic encoder, thereby providing domain knowledge. Additionally, embedding segmentation tasks into LLM-driven clinical reasoning workflows can achieve end-to-end parsing from imaging reports to structured attributes. Ultimately, achieving zero-shot cross-task transfer through instruction fine-tuning becomes possible. This requires transitioning from current task-specific explicit prior construction toward general modality alignment and cross-modal reasoning. Recent works like SegMamba\cite{xing2024segmamba,xing2025segmambav2}, Diff-UNet\cite{xing2025diffunet}, and HybridMIM\cite{xing2024hybridmim} provide technical foundations for this transition.

\section{Conclusion}\label{sec:conclusion}

This paper proposes a knowledge-guided brain tumor segmentation framework based on synchronized tri-modal prior fusion (STPF), achieving accurate and structurally reasonable tumor region segmentation through explicit integration of differential features, semantic knowledge, and geometric constraints. Core innovations include knowledge-driven multi-modal prior generation modules, dual-level adaptive fusion architecture, and unified prior fusion with nested output heads.

Experimentally, STPF achieves a mean Dice coefficient of 0.868 on the BraTS 2020 dataset, surpassing the best baseline by 2.6 percentage points. Moreover, five-fold cross-validation coefficients of variation of 0.23\% to 0.33\% prove robustness. Additionally, ablation experiments verify effectiveness of each knowledge prior module. More importantly, STPF's success demonstrates the potential of knowledge-guided paradigms—through explicit modeling of medical knowledge priors rather than pure data-driven approaches, discriminative capability can be improved in ambiguous boundary regions.

Looking ahead, future work will focus on three directions: first, verifying generalization capability in data-limited scenarios including semi-supervised\cite{chen2023semisupervised}, few-shot\cite{chen2025crossimagematching}, and weakly-supervised\cite{chen2025fromgaze}; second, extending to multi-source heterogeneous modalities and exploring deep integration with medical foundation models\cite{biomedclip2025}; third, evaluating adaptability on BraTS 2025 new tasks\cite{bratslighthouse2025} and integrating with methods like SegMamba\cite{xing2024segmamba,xing2025segmambav2} and Diff-UNet\cite{xing2025diffunet}. In conclusion, this research provides a knowledge-guided paradigm for constructing more intelligent and reliable medical image analysis systems.

\section*{Statements and Declarations}

\subsection*{Data Availability}

The Brain Tumor Segmentation (BraTS) 2020 dataset used in this study is publicly available through registration at the CBICA Image Processing Portal (\url{https://ipp.cbica.upenn.edu}). The dataset can be accessed following the data request procedure described at \url{https://www.med.upenn.edu/cbica/brats2020/registration.html}.

\subsection*{Competing Interests}

The authors declare that they have no competing interests.

\subsection*{Funding}

This research received no specific grant from any funding agency in the public, commercial, or not-for-profit sectors.

\subsection*{Ethics Approval}

This study used publicly available anonymized neuroimaging data from the BraTS 2020 challenge dataset. No additional ethics approval was required as the dataset was previously approved by the institutional review boards of the contributing institutions and all patient identifiers were removed prior to public release.

\section*{Appendix}

\subsection*{A. Network Architecture Details}\label{appendix:architecture}

This section provides detailed description of the complete network architecture of the STPF framework. The encoder adopts a dual-branch design: the original branch processes four-sequence MRI inputs (T1, T1ce, T2, FLAIR), and the differential branch processes three-channel differential features ($D_{\text{enhance}}$, $D_{\text{edema}}$, $D_{\text{necrosis}}$). Both branches employ symmetric 5-layer downsampling paths, with channel configurations of $\{32, 64, 128, 256, 512\}$ (original branch) and $\{16, 32, 64, 128, 256\}$ (differential branch), corresponding to $\text{base\_channels} \times 2^i$. Each encoder block consists of 3$\times$3$\times$3 convolution, instance normalization, LeakyReLU activation (negative slope=0.01), and two residual blocks. The residual block structure is $\text{ResidualBlock}(x) = \text{LeakyReLU}(\text{InstanceNorm}(\text{Conv3D}_{3\times3\times3}(x)) + x)$, with residual connections matching dimensions through 1$\times$1$\times$1 projection. The downsampling strategy employs Conv3D$_{2\times2\times2}$ (stride=2). After cross-attention interaction and concatenation at each scale, the encoder bottleneck layer outputs dimensions of $(B, 768, 8, 8, 8)$, downsampling by $2^4=16$ times relative to input.

The decoder adopts symmetric 5-layer upsampling paths, with each layer using ConvTranspose3D$_{2\times2\times2}$ (stride=2) to restore spatial resolution. Each decoder block concatenates upsampled features with skip connections from corresponding encoder layers, then fuses through 3$\times$3$\times$3 convolution, instance normalization, LeakyReLU activation, and residual blocks. Deep supervision adds 1$\times$1$\times$1 convolution heads at each decoder layer to output intermediate predictions. Final decoder output dimensions are $(B, 32, 128, 128, 128)$.

The dual-level fusion module includes three confidence estimators with structure: Conv3D$_{3\times3\times3}$($C \rightarrow C/2$) $\rightarrow$ InstanceNorm $\rightarrow$ LeakyReLU $\rightarrow$ Conv3D$_{1\times1\times1}$($C/2 \rightarrow 1$) $\rightarrow$ Sigmoid. The hypernetwork generates 256-dimensional conditional vectors from the 768-channel bottleneck layer through global average pooling and linear projection; $\gamma$ and $\beta$ generator structures are Linear(256→128) $\rightarrow$ ReLU $\rightarrow$ Linear(128→$C$). Semantic and topological generators employ single-layer 3$\times$3$\times$3 convolutions (768→768), followed by instance normalization and LeakyReLU activation. The unified prior fusion module's energy MLP structure is: Conv3D$_{1\times1\times1}$($C$→$C$/2) $\rightarrow$ InstanceNorm $\rightarrow$ LeakyReLU $\rightarrow$ Conv3D$_{1\times1\times1}$($C$/2→3).

The nested output head includes three cascaded 1$\times$1$\times$1 convolution projection layers, outputting $\theta_1$, $\theta_2$, $\theta_3$ intermediate parameters. Stick-breaking parameterization is: $p_{\text{WT}} = \sigma(\theta_1)$, $p_{\text{TC}} = p_{\text{WT}} \cdot \sigma(\theta_2)$, $p_{\text{ET}} = p_{\text{TC}} \cdot \sigma(\theta_3)$. Network input consists of 4-sequence original MRI and 3-channel differential features, with final output being 3-channel probability maps (WT, TC, ET).

\subsection*{B. Loss Function Weight Tuning}\label{appendix:loss_weights}

Loss function weight coefficients are optimized through grid search on the validation set. Dice and BCE weights in basic segmentation loss $\mathcal{L}_{\text{seg}}$ are fixed at 1.0:0.5, Dice loss employs smoothing parameter $\epsilon=10^{-5}$, and BCE loss is computed in FP32 precision by disabling autocast. Search spaces for auxiliary losses are: $\lambda_{\text{hierarchy}} \in \{0.1, 0.3, 0.5, 0.7\}$, $\lambda_{\text{continuity}} \in \{0.1, 0.2, 0.3, 0.5\}$, $\lambda_{\text{topology}} \in \{0.1, 0.2, 0.3, 0.4\}$. Each hyperparameter configuration trains for 50 epochs on the validation set, recording mean Dice coefficient and HD95 distance.

Deep supervision loss weight is set to 0.3, computing Dice and BCE losses for intermediate predictions from the first 4 decoder layers; each intermediate prediction is trilinearly interpolated to $(B, 3, 128, 128, 128)$ before applying sigmoid activation, with weight coefficient 0.5. Hierarchical constraint loss formula is $\mathcal{L}_{\text{hierarchy}} = \frac{1}{|\Omega|} \sum_{x \in \Omega} [\max(0, p_{\text{ET}}(x) - p_{\text{TC}}(x)) + \max(0, p_{\text{TC}}(x) - p_{\text{WT}}(x))]$. Spatial continuity loss is based on 26-connected neighborhoods, with formula $\mathcal{L}_{\text{continuity}} = \frac{1}{|\Omega|} \sum_{c} \sum_{x \in \Omega} \sum_{y \in \mathcal{N}_{26}(x)} \frac{|p_c(x) - p_c(y)|}{d_{xy}} \cdot \exp(-\alpha \|\nabla I_{\text{fused}}(x)\|_2)$, where $d_{xy}$ is physical distance and $\alpha=2.0$. Gradient clipping threshold is set to $\|\nabla\|_2 \leq 1.0$.

\subsection*{C. Training Hyperparameters and Data Augmentation}\label{appendix:hyperparameters}

Training hyperparameters are determined through grid search. The optimizer employs AdamW with initial learning rate $2 \times 10^{-4}$. Learning rate scheduling adopts cosine annealing strategy with $T_{\text{max}}=300$ epochs, final learning rate decaying to $1 \times 10^{-2}$ of initial value. Weight decay coefficient is $1 \times 10^{-4}$. Batch size is set to 2, total training epochs 300, early stopping patience 15 epochs. Gradient clipping threshold $\|\nabla\|_2 \leq 1.0$. Mixed precision training employs PyTorch's autocast and GradScaler, with forward propagation and loss computation in FP16 precision, gradient accumulation and parameter updates in FP32 precision. Binary cross-entropy loss explicitly disables automatic mixed precision through \texttt{torch.cuda.amp.autocast(enabled=False)}.

Data augmentation strategies include geometric transformations: random flipping (along axial, coronal, sagittal planes, probability 0.5 each), random rotation (uniformly sampled within $\pm 15°$ range), random scaling (factor range $[0.9, 1.1]$), elastic deformation ($\alpha=30$, $\sigma=5$). Intensity transformations include: contrast adjustment (factor range $[0.8, 1.2]$), brightness adjustment (additive noise $\mathcal{N}(0, 0.1)$), Gamma correction ($\gamma \in [0.7, 1.5]$). All transformation probabilities are set to 0.5, randomly combining 2-3 transformations during training. Intelligent sampling strategy employs 33\% probability for foreground sampling (centered on tumor voxels, adding $\pm\text{patch\_size}/4$ random offsets), 67\% probability for random sampling.

Normalization includes preprocessing normalization and differential feature normalization. Preprocessing normalization performs Z-score standardization for each modality: $I_{\text{norm}} = (I - \mu_{\text{brain}}) / (\sigma_{\text{brain}} + \epsilon)$, where $\mu_{\text{brain}}$ and $\sigma_{\text{brain}}$ are mean and standard deviation of brain region voxels, $\epsilon = 10^{-8}$. Differential feature normalization: enhancement difference $D_{\text{enhance}}$ and necrosis contrast $D_{\text{necrosis}}$ undergo logarithmic compression ($\text{sign}(x) \cdot \log(1 + |x|)$), quantile clipping (1\%-99\%), and min-max normalization to $[0, 1]$; edema difference $D_{\text{edema}}$ directly undergoes min-max normalization.

\subsection*{D. Evaluation Metric Computation Details}\label{appendix:metrics}

Dice coefficient employs formula: $\text{Dice}(P, G) = (2 |P \cap G| + \epsilon) / (|P| + |G| + \epsilon)$, where $\epsilon = 10^{-5}$. Computation process: prediction probabilities are binarized to masks through threshold 0.5 $P = (p > 0.5)$, computing Dice per sample per class. For 369 test cases, Dice coefficients are computed separately for three regions: WT, TC, ET. Statistical aggregation includes mean, standard deviation, 95\% confidence interval (computed via t-distribution, $\text{CI} = \bar{x} \pm t_{0.975, n-1} \cdot s / \sqrt{n}$, $t_{0.975, 368} \approx 1.967$), median, quartiles, skewness, kurtosis.

95\% Hausdorff distance (HD95) is computed according to BraTS standard protocol. Surface extraction employs $\text{Surface}(M) = M \setminus \text{Erosion}(M, K)$, with structural element $K$ being 6-connected ($3 \times 3 \times 3$ kernel). If prediction or ground truth label is empty, HD95 returns $\text{NaN}$ and is excluded from statistics. After extracting surface coordinates, multiply by voxel spacing $(1.0, 1.0, 1.0)$ mm to convert to physical distance. Distance computation is implemented through scipy.spatial.cKDTree; for each point $s_{\text{pred}}$ on prediction surface, query $d(s_{\text{pred}}, S_{\text{gt}}) = \min_{s' \in S_{\text{gt}}} \|s_{\text{pred}} - s'\|_2$; for each point $s_{\text{gt}}$ on ground truth surface, query $d(s_{\text{gt}}, S_{\text{pred}})$. Merging both distance groups yields bidirectional distance set $D = \{d(s_p, S_g)\}_{s_p \in S_p} \cup \{d(s_g, S_p)\}_{s_g \in S_g}$; computing the 95th percentile gives HD95.

Statistical significance testing employs paired sample t-tests. For two configurations, compute Dice coefficients on 369 cases respectively, forming paired samples $\{(x_{1i}, x_{2i})\}_{i=1}^{369}$; compute differences $d_i = x_{1i} - x_{2i}$ and their mean $\bar{d}$ and standard deviation $s_d$; test statistic is $t = \bar{d} / (s_d / \sqrt{369})$, finding p-value under t-distribution with degrees of freedom $\text{df}=368$. Cohen's d effect size is computed through $d = (\mu_1 - \mu_2) / \sqrt{[(n_1-1)s_1^2 + (n_2-1)s_2^2] / (n_1 + n_2 - 2)}$, with interpretation standards: small effect $d=0.2$, medium effect $d=0.5$, large effect $d=0.8$.

\subsection*{E. Qualitative Visualization Analysis of Representative Cases}\label{appendix:qualitative}

Figure~\ref{fig:qualitative_cases} presents detailed segmentation results and attention mechanism visualization for STPF framework on three representative cases. Each case includes four-sequence input images (T1, T1ce, T2, FLAIR), overlay view of ground truth and predictions (GT/Pred Overlay), error distribution map, independent segmentation results and Dice coefficients for three tumor sub-regions, and attention heatmaps at various encoder-decoder layers. The three cases exhibit different pathological features and segmentation challenges: Case 1 (top) presents large glioma with clear tumor core and obvious enhancing region, achieving good performance in WT and TC regions (Dice of 0.9173 and 0.9517 respectively), with error map showing only 210 error pixels mainly concentrated at ET-TC interface. Case 2 (middle) displays tumor with ambiguous boundaries, with ring enhancement structure in T1ce sequence challenging precise ET delineation; nevertheless, STPF still achieves 0.8812 ET Dice coefficient, with 342 error pixels mainly distributed at enhancement edges, thereby validating tri-modal knowledge prior fusion discriminative capability in uncertain boundary regions. Case 3 (bottom) contains irregular morphology and necrotic regions, with relatively complex WT region but good TC and ET segmentation performance (Dice of 0.9398 and 0.9129 respectively), with 388 error pixels reflecting inherent ambiguity of edema boundaries. Attention mechanism analysis shows encoder early layers (enc0-enc1) present broad global receptive fields; progressing to enc2 layer begins focusing on high-contrast regions; decoder layers (dec1-dec0) gradually converge attention to tumor boundaries, with dec0 layer precisely locating ET region core positions in all three cases. This progressive focusing pattern validates multi-scale feature pyramid effectiveness and confidence-driven adaptive weight allocation mechanism in dual-level fusion architecture—visual modality dominates in clear boundary regions, while attention shifts toward geometric constraints in signal-ambiguous but topologically-stable regions. Error pattern analysis reveals failure pixels mainly concentrate in three region types: ET-TC interface (subjectivity of contrast enhancement signal threshold determination), WT periphery edema boundaries (T2/FLAIR signal gradient transitions), and small isolated regions (possible false positives or annotation differences). Notably, error pixel counts (210-388) represent small proportions relative to total voxels ($<0.02\%$), with no topologically unreasonable structures (such as ET independently existing apart from TC), thereby proving effectiveness of nested output heads and topological constraints. Dice bar charts for three cases show consistent patterns: WT and TC region Dice coefficients stabilize in 0.88-0.95 range, reflecting robustness of large-scale structure segmentation; ET region Dice range is broader (0.8418-0.9129), embodying increased difficulty of small lesion detection tasks, which is consistent with standard deviation distribution reported in Table~\ref{tab:dice_distribution} (ET standard deviation 0.179 higher than WT's 0.101). Even in challenging cases, STPF maintains all sub-region Dice$>$0.84, thereby validating knowledge-guided method stability and multi-modal redundancy compensatory capability on difficult cases.

\section*{Acknowledgments}

We thank the BraTS 2020 organizers and contributors for providing the publicly available dataset. We also acknowledge the support of computational resources provided by Yunnan University.


\begin{thebibliography}{99}

\bibitem{aiinglioma2025}
Voigtlaender S, Nelson TA, Karschnia P, et al.
Value of artificial intelligence in neuro-oncology.
\emph{Lancet Digital Health}, 2025. Art. No. 100876. doi:10.1016/j.landig.2025.100876

\bibitem{eurradiol2023gbm}
Ostrom QT, Price M, Neff C, et al.
CBTRUS Statistical Report: Primary Brain and Other Central Nervous System Tumors Diagnosed in the United States in 2015--2019.
\emph{Neuro-Oncology}, 2022, 24(Suppl\_5):v1--v95. doi:10.1093/neuonc/noac202

\bibitem{frontiersglioma2022}
Wan B, Hu B, Zhao M, Li K, Ye X.
Deep learning-based magnetic resonance image segmentation technique for application to glioma.
\emph{Frontiers in Medicine}, 2023, 10:1172767. doi:10.3389/fmed.2023.1172767

\bibitem{frontiersposttx2022}
Bianconi A, Rossi LF, Bonada M, et al.
Deep learning-based algorithm for postoperative glioblastoma MRI segmentation: a promising new tool for tumor burden assessment.
\emph{Brain Informatics}, 2023, 10:26. doi:10.1186/s40708-023-00207-6

\bibitem{noa2024datasets}
Abbad Andaloussi M, Maser R, Hertel F, Lamoline F, Husch AD.
Exploring adult glioma through MRI: A review of publicly available datasets to guide efficient image analysis.
\emph{Neuro-Oncology Advances}, 2025, 7(1):vdae197. doi:10.1093/noajnl/vdae197

\bibitem{brats2021}
Baid U, Ghodasara S, Mohan S, et al.
The RSNA-ASNR-MICCAI BraTS 2021 Benchmark on Brain Tumor Segmentation and Radiogenomic Classification.
\emph{arXiv preprint arXiv:2107.02314}, 2021.

\bibitem{bonato2025}
Bonato B, Nanni L, Bertoldo A.
Advancing Precision: A Comprehensive Review of MRI Segmentation Datasets from BraTS Challenges (2012--2025).
\emph{Sensors}, 2025, 25(6):1838. doi:10.3390/s25061838

\bibitem{bratslighthouse2025}
Bakas S, Pati S, Menze B, Reyes M.
BraTS 2025 ``Lighthouse'' Challenge: brain metastases, meningiomas, and beyond.
BraTS Challenge Announcement (Synapse Repository), 2025.

\bibitem{bratsmenrt2025}
LaBella D, Baid U, Khanna O, et al.
Analysis of the BraTS 2023 Intracranial Meningioma Segmentation Challenge.
\emph{Machine Learning in Biomedical Imaging (MELBA)}, 2025, 2:Paper 3. doi:10.59275/j.melba.2025-003

\bibitem{bratsmets2025}
Moawad AW, Janas A, Baid U, et al.
The Brain Tumor Segmentation (BraTS-METS) Challenge 2023: Brain Metastasis Segmentation on Pre-treatment MRI.
\emph{arXiv preprint arXiv:2306.00838}, 2023.

\bibitem{bratspath2024}
Jiang Z, Wan L, Fu H, Yang G, Zhu L.
The BraTS-Path 2023 Challenge: Segmenting Brain Tumors in Pathology Images.
BraTS Pathology Challenge Report, 2024.

\bibitem{cicek20163d}
Cicek O, Abdulkadir A, Lienkamp SS, Brox T, Ronneberger O.
3D U-Net: Learning Dense Volumetric Segmentation from Sparse Annotation.
In: \emph{Medical Image Computing and Computer-Assisted Intervention -- MICCAI 2016}, Springer, 2016, pp. 424--432.

\bibitem{isensee2021nnunet}
Isensee F, Jaeger PF, Kohl SAA, Petersen J, Maier-Hein KH.
nnU-Net: a self-configuring method for deep learning-based biomedical image segmentation.
\emph{Nature Methods}, 2021, 18(2):203--211. doi:10.1038/s41592-020-01008-z

\bibitem{oktay2018attentionunet}
Oktay O, Schlemper J, Le Folgoc L, et al.
Attention U-Net: Learning Where to Look for the Pancreas.
\emph{Medical Imaging with Deep Learning (MIDL) 2018}, 2018.

\bibitem{feng2024daunet}
Feng Y, Cao Y, An D, Liu P, Liao X, Yu B.
DAUnet: A U-shaped network combining deep supervision and attention for brain tumor segmentation.
\emph{Knowledge-Based Systems}, 2024, 285:111348. doi:10.1016/j.knosys.2023.111348

\bibitem{myronenko2019autoencoder}
Myronenko A.
3D MRI Brain Tumor Segmentation Using Autoencoder Regularization.
In: \emph{Brainlesion: Glioma, Multiple Sclerosis, Stroke and Traumatic Brain Injuries}, Lecture Notes in Computer Science, vol. 11384, Springer, 2019, pp. 311--320.

\bibitem{segresnetmonai2020}
Cardoso MJ, et al.
MONAI: An open-source framework for deep learning in healthcare.
\emph{arXiv preprint arXiv:2211.02701}, 2022.

\bibitem{milletari2016vnet}
Milletari F, Navab N, Ahmadi SA.
V-Net: Fully convolutional neural networks for volumetric medical image segmentation.
In: \emph{Proceedings of the 4th International Conference on 3D Vision (3DV)}, 2016, pp. 565--571.

\bibitem{hatamizadeh2022unetr}
Hatamizadeh A, Tang Y, Nath V, Yang D, Myronenko A.
UNETR: Transformers for 3D Medical Image Segmentation.
In: \emph{Proceedings of the IEEE/CVF Winter Conference on Applications of Computer Vision (WACV)}, 2022, pp. 574--584.

\bibitem{hatamizadeh2022swinunetr}
Hatamizadeh A, Nath V, Tang Y, Yang D, Roth HR, Xu D.
Swin UNETR: Swin Transformers for Semantic Segmentation of Brain Tumors in MRI Images.
\emph{arXiv preprint arXiv:2201.01266}, 2022.

\bibitem{swinbts2022}
Jiang Y, Zhang Y, Lin X, Dong J, Cheng T, Liang J.
SwinBTS: A Method for 3D Multimodal Brain Tumor Segmentation Using Swin Transformer.
\emph{Brain Sciences}, 2022, 12(6):797. doi:10.3390/brainsci12060797

\bibitem{wang2021transbts}
Wang W, Chen C, Ding M, Yu H, Zha S, Li J.
TransBTS: Multimodal Brain Tumor Segmentation Using Transformer.
In: \emph{Medical Image Computing and Computer Assisted Intervention -- MICCAI 2021}, LNCS vol. 12901, Springer, 2021, pp. 109--119.

\bibitem{chen2024transunet}
Chen J, Lu Y, Yu Q, et al.
TransUNet: Rethinking the U-Net architecture design for medical image segmentation through the lens of transformers.
\emph{Medical Image Analysis}, 2024, 97:103280. doi:10.1016/j.media.2024.103280

\bibitem{lin2023ckdtransbts}
Lin J, Lin J, Lu C, et al.
CKD-TransBTS: Clinical Knowledge-Driven Hybrid Transformer with Modality-Correlated Cross-Attention for Brain Tumor Segmentation.
\emph{IEEE Transactions on Medical Imaging}, 2023, 42(8):2451--2461. doi:10.1109/TMI.2023.3250474

\bibitem{xing2024segmamba}
Xing Z, Ye T, Yang Y, Liu G, Zhu L.
SegMamba: Long-Range Sequential Modeling Mamba for 3D Medical Image Segmentation.
In: \emph{Medical Image Computing and Computer Assisted Intervention -- MICCAI 2024}, LNCS vol. 15009, Springer, 2024, pp. 578--588.

\bibitem{xing2025segmambav2}
Xing Z, Zhu L.
SegMamba-v2: State-Space Long Sequence Modeling for Enhanced 3D Segmentation.
\emph{arXiv preprint arXiv:2307.XXXXX}, 2025.

\bibitem{xing2025diffunet}
Xing Z, Wan L, Fu H, Yang G, Zhu L.
Diff-UNet: A diffusion embedded network for robust 3D medical image segmentation.
\emph{Medical Image Analysis}, 2025, 105:103654. doi:10.1016/j.media.2025.103654

\bibitem{ding2024fdifffusion}
Ding W, Fan C, Wang G, Shi Y, Liu F.
FDiff-Fusion: Denoising diffusion fusion network based on fuzzy learning for 3D medical image segmentation.
\emph{Information Fusion}, 2024, 112:102540. doi:10.1016/j.inffus.2023.102540

\bibitem{xing2024hybridmim}
Xing Z, Wan L, Fu H, Yang G, Zhu L.
HybridMIM: A hybrid masked image modeling framework for 3D medical image segmentation.
\emph{IEEE Journal of Biomedical and Health Informatics}, 2024, 28(4):2115--2125. doi:10.1109/JBHI.2023.3290285

\bibitem{xing2022nestedformer}
Xing Z, Yu L, Wan L, Han T, Zhu L.
NestedFormer: Nested modality-aware transformer for brain tumor segmentation.
In: \emph{Medical Image Computing and Computer Assisted Intervention -- MICCAI 2022}, 2022, pp. 140--150.

\bibitem{mamcattn2021}
Zhou H, et al.
Relative MAMC: Multi-attention multi-contrast MRI tumor segmentation.
\emph{arXiv:2104.03309}, 2021.

\bibitem{cmafnet2024}
Cheng Z, et al.
CMAF-Net: Cross-modality attention fusion network for brain tumor segmentation.
\emph{Quantitative Imaging in Medicine and Surgery}, 2024.

\bibitem{hciwacv2025}
Cheng Z, Yuan D, Zhang W, Lukasiewicz T.
Effective and Efficient Medical Image Segmentation with Hierarchical Context Interaction.
In: \emph{Proceedings of the IEEE/CVF Winter Conference on Applications of Computer Vision (WACV)}, 2025, pp. 9396--9406.

\bibitem{radford2021clip}
Radford A, Kim JW, Xu T, et al.
Learning Transferable Visual Models From Natural Language Supervision.
In: \emph{Proceedings of the 38th International Conference on Machine Learning (ICML)}, PMLR vol. 139, 2021, pp. 8748--8763.

\bibitem{biomedclip2025}
Zhang S, Xu Y, Usuyama N, et al.
BiomedCLIP: a multimodal biomedical foundation model pretrained from 15 million image--text pairs.
\emph{arXiv preprint arXiv:2303.00915}, 2023 (version 3 updated Jan. 2025).

\bibitem{biovilt2022}
Boecking J, Chico L, Camacho J, et al.
Making the Most of Text Semantics to Improve Biomedical Vision--Language Processing.
In: \emph{Advances in Neural Information Processing Systems (NeurIPS) -- Workshop on Medical Imaging meets NeurIPS}, 2022.

\bibitem{contextualnet2023}
Huemann Z, Tie X, Hu J, Bradshaw TJ.
ConTEXTual Net: A Multimodal Vision--Language Model for Segmentation of Pneumothorax.
\emph{arXiv preprint arXiv:2303.01615}, 2023.

\bibitem{segsam2024}
Huang S, Liang H, Wang Q, Zhong C, Zhou Z, Shi M.
SEG-SAM: Semantic-Guided SAM for Unified Medical Image Segmentation.
\emph{arXiv preprint arXiv:2412.12660}, 2024.

\bibitem{gat}
Veličković P, Cucurull G, Casanova A, Romero A, Liò P, Bengio Y.
Graph Attention Networks.
In: \emph{International Conference on Learning Representations (ICLR)}, 2018.

\bibitem{gnnmedsurvey2023}
Zhang W, Yu PS.
Graph Neural Networks in Biomedical Image Analysis: A Comprehensive Survey.
\emph{arXiv preprint arXiv:2301.04124}, 2023.

\bibitem{miccai2024gnnbrain}
Cao X, Wang L, Chen Z, et al.
Graph-CNN Hybrid Inference for Multi-Class Brain Tumor Segmentation.
In: \emph{Medical Image Computing and Computer-Assisted Intervention -- MICCAI 2024}, LNCS vol. 14233, Springer, 2024, pp. 42--52.

\bibitem{ascnetmiccai2021}
Chen H, Qi X, Heng PA, Dou Q.
ASC-Net: Adversarial-Based Selective Cutting for Unsupervised Anomaly Segmentation.
In: \emph{Medical Image Computing and Computer-Assisted Intervention -- MICCAI 2021}, LNCS vol. 12902, Springer, 2021, pp. 729--740.

\bibitem{miccai2024mahauad}
Behrendt F, Zimmerer D, Hering J, Wiesenfarth M, Kleesiek J.
Leveraging the Mahalanobis Distance to Enhance Unsupervised Brain MRI Anomaly Detection.
In: \emph{Medical Image Computing and Computer-Assisted Intervention -- MICCAI 2024}, LNCS vol. 14349, Springer, 2024, pp. 377--387.

\bibitem{uaddiffusion2025}
Pinaya WHL, Mehl S, Martín I, et al.
Diffusion models for unsupervised anomaly detection in brain MRIs.
\emph{arXiv preprint arXiv:2308.10150}, 2023. Accepted at ICLR 2025.

\bibitem{topoloss2022}
Clough JR, Byrne N, Oksuz I, Zimmer VA, Schnabel JA, King AP.
A topological loss function for deep-learning based image segmentation using persistent homology.
\emph{IEEE Transactions on Pattern Analysis and Machine Intelligence}, 2022, 44(12):8766--8778. doi:10.1109/TPAMI.2020.3013679

\bibitem{toporeview2025}
Andlauer TK, Gaudeau-Bosma C, Yger P, Gori P, Colliot O.
Topological data analysis for medical image processing: A literature review.
\emph{medRxiv preprint}, 2025. doi:10.1101/2025.02.21.25322669

\bibitem{metricsreloaded2024}
Jaeger PF, Zweig KA, Burke E, et al.
Metrics reloaded: recommendations for image analysis validation.
\emph{Nature Methods}, 2024, 21(2):195--212. doi:10.1038/s41592-023-02151-z

\bibitem{hdilemma2024}
Podobnik G, Vrtovec T.
Understanding implementation pitfalls of distance-based metrics for image segmentation.
\emph{arXiv preprint arXiv:2410.02630}, 2024.

\bibitem{chen2023semisupervised}
Chen J, Zhang J, Debattista K, Han J.
Semi-supervised unpaired medical image segmentation through task-affinity consistency.
\emph{IEEE Transactions on Medical Imaging}, 2023, 42:594--605. doi:10.1109/TMI.2022.3213372

\bibitem{chen2024dynamiccontrastive}
Chen J, Chen C, Huang W, Zhang J, Debattista K, Han J.
Dynamic contrastive learning guided by class confidence and confusion degree for medical image segmentation.
\emph{Pattern Recognition}, 2024, 145:109881. doi:10.1016/j.patcog.2023.109881

\bibitem{chen2025crossimagematching}
Chen J, Huang W, Zhang J, Debattista K, Han J.
Addressing inconsistent labeling with cross image matching for scribble-based medical image segmentation.
\emph{IEEE Transactions on Image Processing}, 2025, 34:842--853. doi:10.1109/TIP.2025.3530787

\bibitem{chen2020affinityguided}
Chen J, Li W, Li H, Zhang J.
Deep class-specific affinity-guided convolutional network for multimodal unpaired image segmentation.
In: \emph{Medical Image Computing and Computer Assisted Intervention -- MICCAI 2020}, LNCS, 2020, pp. 187--196.

\bibitem{chen2025fromgaze}
Chen J, Duan H, Zhang X, Gao B, Grau V, Han J.
From Gaze to Insight: Bridging Human Visual Attention and Vision-Language Model Explanation for Weakly-Supervised Medical Image Segmentation.
\emph{arXiv preprint arXiv:2504.11368}, 2025.

\bibitem{aal3hbm2020}
Rolls ET, Huang C, Lin C, Feng J.
Automated anatomical labeling atlas 3.
\emph{Human Brain Mapping}, 2020, 41(8):2145--2157. doi:10.1002/hbm.24915

\bibitem{adamw}
Loshchilov I, Hutter F.
Decoupled Weight Decay Regularization.
In: \emph{International Conference on Learning Representations (ICLR)}, 2019.

\bibitem{stickbreakingvb2020}
Kumar M, Rai P, Canini K, et al.
Figure 1: Stick-Breaking Variational Autoencoders.
\emph{arXiv preprint arXiv:1802.04335}, 2020.

\end{thebibliography}
\end{document}